\documentclass[lettersize,journal]{IEEEtran}
\usepackage{amsmath,amsfonts}
\usepackage{algorithmic}
\usepackage{algorithm}
\usepackage{array}
\usepackage{textcomp}
\usepackage{stfloats}
\usepackage{url}
\usepackage{verbatim}
\usepackage{graphicx}
\usepackage{cite}
\graphicspath{{img/}}
\usepackage{amssymb}
\usepackage{booktabs}        
\usepackage{amsfonts}        
\usepackage{amsmath}         
\usepackage{tabularx}        
\usepackage{multirow}        
\usepackage{url}             
\usepackage[numbers]{natbib} 
\usepackage[dvipsnames]{xcolor}
\usepackage[pdftex]{hyperref}
\hypersetup{
   colorlinks,
   linkcolor={red},
   citecolor={black},
   urlcolor={black}
}
\usepackage{tikz}
\usepackage{siunitx}
\usetikzlibrary{arrows, calc, external, fit, positioning, shapes, spy}
\usepackage{subcaption}
\usepackage{wrapfig}

\definecolor{arrowBlue}{RGB}{70, 113, 211}
\definecolor{arrowGreen}{RGB}{141, 211, 95}
\definecolor{arrowRed}{RGB}{138, 0, 0}
\definecolor{blueDiagram12}{RGB}{0, 132, 209}

\newcommand{\estimates}{\overset{\scriptscriptstyle\wedge}{=}}

\usepackage{fontawesome5}
\newcommand{\ORCID}[1]{\thinspace\textsuperscript{\href{https://orcid.org/#1}{\textcolor[HTML]{A6CE39}{\faOrcid}}}}
\newcommand{\ORCIDBeuth}{0000-0001-5482-9787}     
\newcommand{\ORCIDKowerko}{0000-0002-4538-7814}   

\begin{document}

\bstctlcite{IEEEexample:BSTcontrol}

\title{On the Transfer of Collinearity to Computer Vision}

\author{Frederik Beuth\ORCID{\ORCIDBeuth}, Danny Kowerko\ORCID{\ORCIDKowerko}
\thanks{Affilation: Junior Professorship of Media Computing, Chemnitz University of Technology, 09107 Chemnitz, Germany.}%
\thanks{Corresponding author: Frederik Beuth, \textit{E-mail address: frederik.beuth@posteo.de}}%
\vspace*{-1em}
}

\maketitle

\begin{abstract}
  Collinearity is a visual perception phenomenon in the human brain that amplifies spatially aligned edges arranged along a straight line. However, it is vague for which purpose humans might have this principle in the real-world, and its utilization in computer vision and engineering applications even is a largely unexplored field. In this work, our goal is to transfer the collinearity principle to the computer vision domain, and we explore the potential usages of this novel principle for computer vision applications. We developed a prototype model to exemplify the principle, then tested it systematically, and benchmarked it in the context of four use cases. Our experimental use cases are selected to spawn a broad range of potential applications and scenarios: sketching the combination of collinearity with deep learning (case I and II), using collinearity with saliency models (case II), and utilizing collinearity as a feature detector (case I). In the first use case, we found that collinearity is able to improve the fault detection of semiconductor wafers and obtain a performance increase by a factor 1.24 via collinearity (decrease of the error rate from 6.5\,\% to 5.26\,\%). In the second use case, we test the defect recognition in nanotechnology materials and achieve a performance increase by 3.2x via collinearity (deep learning, error rate from 21.65\,\% to 6.64\,\%), and furthermore explore saliency models within the case. As third experiment, we cover occlusions; while as fourth experiment, we test ImageNet and we observe that it might not be very beneficial for ImageNet. 
  Therefore, we are able to assemble a list of scenarios for which collinearity is beneficial (wafers, nanotechnology, occlusions), and for what is not beneficial (ImageNet). Hence, we infer collinearity might be suitable for industry applications: It helps the recognition and detection if the image structures of interest are man-made because such structures often consist of lines or line-elements. By exploring this principle of collinearity, we hope we can provide another tool for the computer vision community to capture the power of human visual processing.
\end{abstract}


\setlength{\textfloatsep}{1.5em}

\section{Introduction}
\label{sec:introduction}

\begin{figure}
    \centering
    \includegraphics[width=0.94\columnwidth]{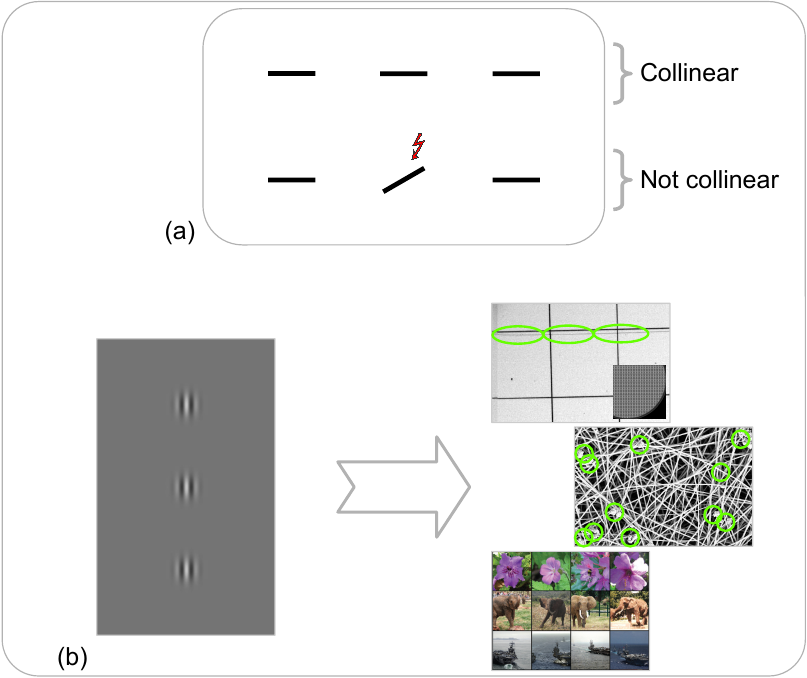}
    \caption{\textbf{a)} Illustration of the collinearity principle. \textbf{b)} Transfer from psychology to the computer vision domain. \textbf{b, left)} Psychology, reprinted from \citep{Maniglia2015a}. \textbf{b, right)} Computer vision domain: Fault detection in semiconductor wafers \citep{Beuth2021}. Defect recognition in nanotechnology materials \citep{Carrera2017}, and ImageNet subset taken from \citep{Krizhevsky2017} (in clock-wise direction). The green circles denote the structures of interests in \citep{Beuth2021,Carrera2017}.
    }
    \label{fig:introduction}
    \vspace*{-0.75em}
\end{figure}

In computer vision, it is a long-term dream to capture the power of the human brain and its processing. For this reason, we try to transfer the power of biological processing principles to computer vision and engineering. One of these principles is collinearity, which amplifies spatially aligned edges.

Collinearity is part of the psychophysical principles of grouping and perceiving objects of human subjects \citep{Schmidt2009,Wagemans2012a}. In humans, edges as well as lines are among others better detected in images with cluttered contents \citep{Polat1993}, are better detected at worse seeing conditions such as low contrasts \citep{Maniglia2015a}, and are generally better perceived. The effect occurs when a set of edges is aligned in a longer line, denoted as collinearity (Fig. \ref{fig:introduction}a) \citep{Maniglia2015a,Polat1993,Kapadia1995,Schmidt2009}. Typically, in such studies, they investigate the perception of a \textit{target} object surrounded by other stimuli in a collinearly aligned arrangement, whereby the other stimuli are denoted as \textit{flankers} (Fig. \ref{fig:introduction}a).

A corpus of psychological studies exists investigating the properties of this principle. The works of Maniglia et al. (2015) \citep{Maniglia2015a} and Polat \& Sagi (1993) \citep{Polat1993} show that only lines aligned at a certain distance are facilitated. Other studies show the effects when the lines are not well oriented to each other, e.g., flankers that are orthogonal together \citep{Maniglia2015a,Polat1993}, and many works outline that the principle facilitates lines that are interrupted (e.g. \citep{Maniglia2015a}). Another work reviewing the concept in general is for example \citep{Polat1994a}. 
The properties of collinearity are fairly complex since its effects depend also on the spatial frequency of the line pattern \citep{Polat2009}, on its eccentricity away from the fovea \citep{Shani2005}, on its embedding in a complex image pattern \citep{Bonneh1998}, and it can be modulated by visual attention \citep{Shani2005}. The investigations of the principle of collinearity date back to the discovery of Gestalt principles in 1912 \citep{Wertheimer1923,Wagemans2012a}. Non-computational models and theories, e.g. from the field of Gestalt- and visual perception, can be found in Wagemans et al. (2012b) \citep{Wagemans2012b}.

As outlined by the aforementioned, large corpus of data, collinearity is a notable principle in the human brain, since it is involved in many behavioral vision paradigms. Therefore, collinearity must be of purpose for something, as it is prominently present in the visual system of humans, and we would like to explore the utilization of collinearity processing here.

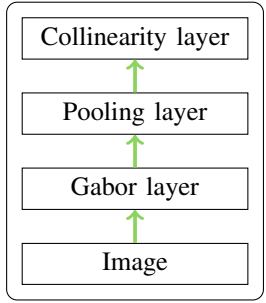
\begin{figure}[t]
	\centering
    \begin{tikzpicture}[
        r/.style={draw, rectangle}]
    \node[r, minimum width=3cm] (node4) at (1, -1) {Collinearity layer};
    \node[r, minimum width=3cm] (node3) at (1, -2) {Pooling layer};
    \node[r, minimum width=3cm] (node2) at (1, -3) {Gabor layer};
    \node[r, minimum width=3cm] (node1) at (1, -4) {Image};
    \node[draw, inner sep=0.2cm, rounded corners, fit=(node1) (node2) (node3) (node4)] (d3) {};
    \draw[<-, line width=0.5mm, arrowGreen] (node2) -- (node1);
    \draw[<-, line width=0.5mm, arrowGreen] (node3) -- (node2);
    \draw[<-, line width=0.5mm, arrowGreen] (node4) -- (node3);
	\end{tikzpicture}
    \caption{Neural model of collinearity.}
    \label{fig:model}
    \vspace*{-0.75em}
\end{figure}

There exist a few previous works regarding models of collinearity. The early work of Krüger (1998) \citep{Kruger1998} have measured the second-order statistics of natural images, formalized them into a statistical model, and pointed out that they underly collinearity. Carreira et al. (1998) \citep{Carreira1998} have developed a neuro-computational model for the domain of computer vision and artificial intelligence. They extracted edge information in scenes via Gabor filters \citep{Jones1987}, and then group the edge features together for the goal of the improved detection of man-made structures, i.e. bridges. The later work by Mundhenk, Nathan \& Itti (2005) \citep{Mundhenk2005} presented a more extensively developed model based on neuro-computational ideas, which investigates collinearity in combination with contour integration. Their aim was to improve contour integration for visual saliency \citep{Itti1998}, hence their work resides more in the fields of saliency models and contour integration. Methodologically, their system utilizes Gabor filters for edge detection, and on top, a neuronal connection pattern that enhances linked edges via interneurons and a complex microcircuit. Some other works address a loosely related topic, the learning of the underlying connectivity. The study of Hoyer \& Hyvärinen (2002) \citep{Hoyer2002} learns the connectivity from natural scenes and utilizes sparse coding in their neuro-computational model, showing that the connectivity supports a complex pattern to realize the edge patterns found in natural images. Another work \citep{Prodohl2003} uses a similar approach to learn the neuronal connectivity, but in the field of object motion and with a Hebbian learning rule. 
Yet, the interest in the field of collinearity seems to have declined after the 2010s as our literature search shows, we suspect that this is due to the rise of deep learning after 2012 since many labs have switched to the novel topic. Nevertheless, this is our motivation to revisit this field of collinearity. 

\textit{Contribution:} The research field of collinearity is well explained in psychology. It is involved in many behavioral paradigms in human vision, so it is highly probably that it must be vital for some purpose. However, its implications as well as usages in applications have only been explored by a very few studies. This motivates us to revisit this field and to analyze the practical usage of collinearity in computer vision and engineering systems. At the end of the paper, our goal is to develop a list of applications for which the principle has advantages and a list of application(s) for which it has disadvantages. Our work is based on a previous, different modeling work of us that simulates small neural circuits \citep{Beuth2015a}, and here we will develop a neural circuit for collinearity to transfer this principle to computer vision.

\newcommand{\Area}[1]{^{\mbox{\tiny #1}}}
\newcommand{\excRF}[2]{\text{excRF}\left(#1,#2\right)}
\newcommand{\RF}[2]{\text{RF}^{#1}_{x'}\left(#2\right)} 
\newcommand{\RFin}[1]{{{x'} \in \text{RF}\left(x,#1\right)}}
\newcommand{\mean}[2]{\text{M}_{#1}\left(#2\right)}
\newcommand{\pos}[1]{\left(#1\right)^+}
\newcommand{\posT}[1]{f_1\left(#1\right)}
\newcommand{\normZeroOne}[1]{f_2\left(#1\right)}
\newcommand{\mdeg}{^{\circ}}

\newcolumntype{L}[1]{>{\raggedright}p{#1}}
\newcolumntype{P}[1]{>{ \centering  \arraybackslash }p{#1}}

\newcolumntype{Q}[1]{>{ \raggedleft \arraybackslash }p{#1}}
\newcolumntype{M}[1]{>{ \centering  \arraybackslash }m{#1}}
\newcolumntype{N}[1]{>{ \raggedleft \arraybackslash }m{#1}}
\newcolumntype{B}[1]{>{ \centering  \arraybackslash }b{#1}}
\newcolumntype{C}[1]{>{ \raggedleft \arraybackslash }b{#1}}

\begin{figure}[t]
    \centering
    \includegraphics[width=1.00\columnwidth]{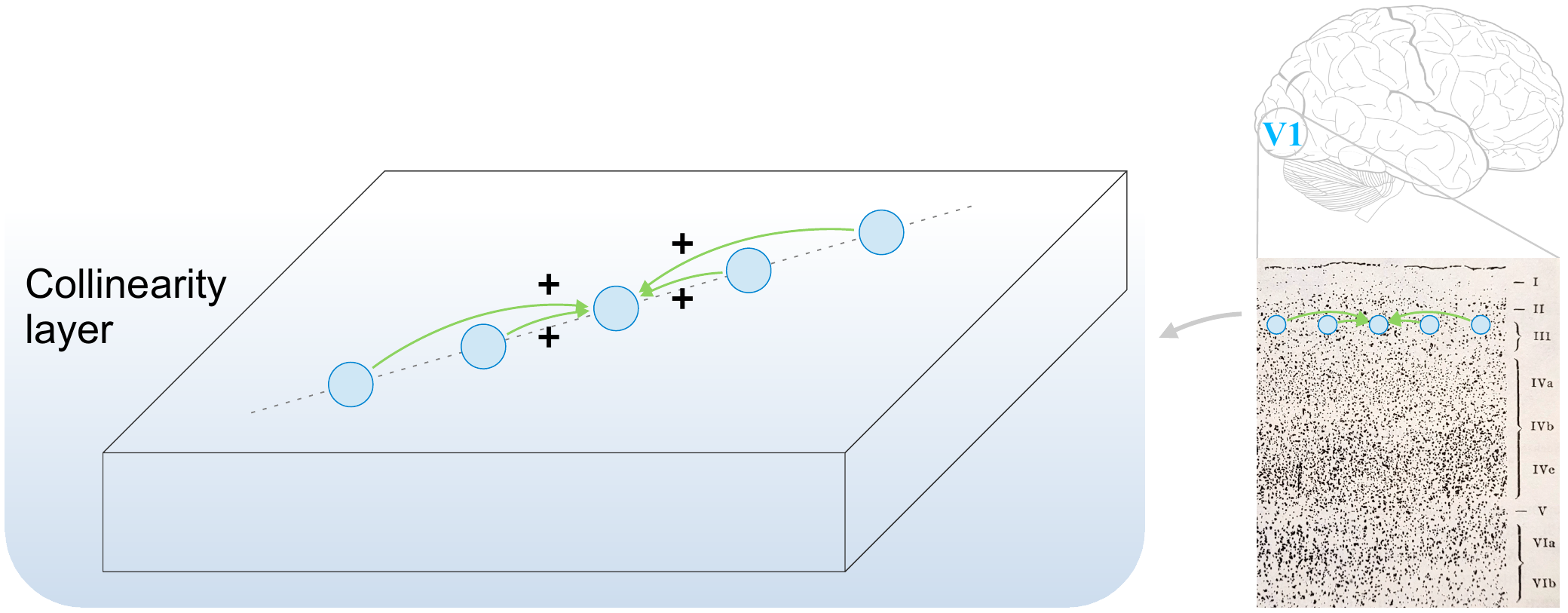}
    \caption{Neural connections which mediate the concept of collinearity in the model. The connections link neurons along a virtual line (collinearity), and their influence enhances the central neuron. They are located inside the collinearity layer. Note, only one of the multiple connection orientations is shown.
    In the right inset, it is shown the assumed location of the model in the human brain (primary visual cortex, V1) and the cortical layers (reprinted from \citep{Brodmann1909}).
    }
    \label{fig:modelConnections}
\end{figure}

\section{Model}

\subsection{Overview}
\label{sec:model}

We have developed a model of collinearity as depicted in Fig. \ref{fig:model}, which consists of a feature extraction stage, a pooling layer, and a collinearity layer. 
The model processes a grayscale or RGB image and filters it via convolution operations with a bank of Gabor filters to detect oriented edges (features). Collinearity is a property commonly tested with edges, thus, an edge-filter layer is implemented, yet more sophisticated features than edges are also surely possible. Gabor filters are a common method for detecting edges \citep{Jones1987}. Note, that we use Gabor filters in pixel space and not frequency space, as they  have been employed in our previous works \citep{Beuth2019,Antonelli2014,Hamker2005b}. These filters were introduced by Jones \& Palmer (1987) \citep{Jones1987} and are commonly utilized to model neural operations in the primate brain for recognizing edges. 
Depending on the model configuration, we use 8 or up to 32 orientations of the edge filters. In the next layer, the neural activities are spatially pooled (pooling layer). 
As a pooling operation, we investigate max-pooling as well as a bilinear image resizing operation, and finally decide due to a better performance for a bilinear image resizing operation with a Lanczos3 kernel \citep{Beuth2019,Turkowski1990} since it is an often-utilized standard method for reducing image resolution. 
The next stage of the model contains the neural connectivity that simulates collinearity (collinearity layer). The connectivity links identically-oriented edges that are arranged in a straight line (Fig. \ref{fig:modelConnections}). The central neuron is influenced by the neurons in its neighborhood, which are arranged on a straight line. The connectivity pattern is required for each orientation of the edges, thus, in our example with eight orientations, this results in eight connectivity patterns. In the proposed model, the influence of collinearity is applied as a multiplicative enhancement (Eq. \ref{eq:col3}), scaling the response up or down. We decided to apply it in such a kind since it is inspired by the finding about visual attention that its signal influence is multiplicative \citep{Beuth2015a}. The collinearity layer is the top-layer of the model, whose neural activity can then be read out or utilized for other detection modules depending on the task at hand.

\subsection{Mathematical equations of the model}
\label{sec:eq}

The model is constituted by the set of the following equations. In them, the indices $\{x_1,\,x_2\}$ denote the 2D image space, while $l$ denotes the feature, i.e. orientation. The term $r_{x_1,\,x_2,\,l}$ denotes the firing rate of a neuron, while all layers are 3D arrays.

\vspace{1em}
\textit{Gabor layer}
\begin{eqnarray}
	\label{eq:gabor}
	r\Area{Gabor}_{x_1,\,x_2,\,l} &=& [r\Area{Image}_{x_1,\,x_2} * G_l]^+ \\
	G_{x_1,\,x_2,\,l} &=& exp\left( -\left( \frac{X_{1,\,l}^2}{2\sigma_1^2} + \frac{X_{2,\,l}^2}{2\sigma_2^2}\right)\right) \cdot \\
	\nonumber && \cos\left(2 \pi f X_{1,\,l} + \psi\right) 
\end{eqnarray}
\vspace*{-2em}
\begin{eqnarray}	
	\nonumber\mbox{with: } \quad\theta_l &=& \left\{0\pi, ~0.25\pi, ~ ... ~ 1.75\pi\right\} \\
	X_{1,\,l} &=& x_1  \cos \theta_i + x_2 \sin \theta_i \\
	X_{2,\,l} &=& -x_1  \sin \theta_i + x_2 \cos \theta_i 
\end{eqnarray}

Whereby $G$ denotes a rotated Gabor kernel \citep{Jones1987}, and $\theta_l$ is its utilized orientation (up to 8, 16 or 32 orientations dependent on the use case). The symbol $*$ denotes convolution and $[~]^+$ half-rectification. We build the kernel in such a way that the free parameter of the kernel is the parameter discretization window size of the kernel, denoted kernel size $K$. By default, it is set to $11 \times 11$ pixels. Hence, the discretization points $x_1$, $x_2$ are implicitly the free parameter(s) and are considered in the range of $-5\le x \le 5$ as default. The parameter K determines the other variables in the equation, $f$, $\sigma_1$, and $\sigma_2$ through choosing them appropriately for a given $K$.

\vspace{1em}
\textit{Pooling layer}
\begin{eqnarray}
	r\Area{Pooling}_{x_1,\,x_2,\,l} &=& r\Area{Gabor}_{x_1,\,x_2,\,l} * H * H^T \\
	\label{eq:lanczos3}H_x &=& \left\{ \begin{array}{ll} 1 & x'=0 \\ \dfrac{a \, \sin(\pi \,x') \, \sin(\pi \,x'/a) } {\pi^2 \, x'^2} & 0 < |x'| < a \\ 0 & x' \ge a \end{array} \right. \\ 
	\nonumber \mbox{with} &:& a=3, \quad x' = x/p
\end{eqnarray}

Whereby for pooling, we use a bilinear image resize operation with a Lanczos3 kernel \citep{Beuth2019,Turkowski1990} (Lanczos filter with lobed parameter $a=3$). The term $H$ denotes the used kernel, here the Lanczos3 kernel \citep{Turkowski1990}. The size of the pooled region is denoted by $p=3$, denoting a pooling over a $3\times3$ area.

\vspace{1em}
\textit{Collinearity layer}
\begin{eqnarray}
	\label{eq:col1}
	\tau\Area{Col} \, \frac{\partial r\Area{Col}_{x_1,\,x_2,\,l}}{\partial t} &=& - r\Area{Col}_{x_1,\,x_2,\,l} + r\Area{Pooling}_{x_1,\,x_2,\,l} \cdot \\
	\nonumber &&\left(1 + w\Area{Col} \cdot a\Area{Col}_{x_1,\,x_2,\,l}\right) \\[0.2em]
	\nonumber\mbox{with: }~ a\Area{Col}_{x_1,\,x_2,\,l} &=& r\Area{Col}_{x_1,\,\,x_2,\,l} * w_{x'_1,\,x'_2,\,l} \label{eq:col3}
\end{eqnarray}

Where $a\Area{Col}$ denotes the collinearity influence, which the current neuron at location $\{x_1,\,x_2\}$ receive from its neighboring neurons at location $\{x'_1,\,x'_2\}$ via the collinearity connections. The variable $w_{x'_1,\,x'_2,\,l}$ represents the collinearity connectivity matrix. The matrix is the core element of the model and its design is outlined in the next section (Sec. \ref{sec:modelNeuroAndKernel}). Spatial extent and width of the neural collinearity connectivity define the size of the matrix, and thus the matrix's indices $x'_1$ and $x'_2$ range over the spatial extent and width appropriately. The connection pattern is also rotated appropriately in the 2D image space according to the neural collinear orientation, and thus the collinearity connectivity is specific for a particular orientation $l$. The symbol $*$ denotes convolution. In the equation \ref{eq:col1}, a scalar $w\Area{Col}$ can be used to scale up or scale down the influence of the collinearity $a\Area{Col}$.

The fire rates of neurons in the collinearity layer are simulated with ordinary differential equations (ODEs) to simulate the parallel influences in the model and to simulate the neural firing rates. It is a similar approach as in our previous works \citep{Beuth2019,Beuth2015a,Hamker2005b} and it is heavily utilized in \citep{Beuth2015a}. The approach solves an ODE iteratively via the Euler method (numerical integration of ODEs).
This type of ordinary differential equation for simulating the neural firing rates converges, after several time steps, to a fixed value that represents the final firing rate. The equation in Eq. \ref{eq:col1} converges to the value $r\Area{Pooling}_{x_1,\,x_2,\,l} \cdot \left(1 + w\Area{Col} \cdot a\Area{Col}_{x_1,\,x_2,\,l}\right)$. With $\tau\Area{Col} = 15 \, ms$.

\subsection{Neurophysiological foundation and design of the collinearity connectivity pattern}
\label{sec:modelNeuroAndKernel}

The core component of the model is the collinearity connectivity matrix. The biological precision of the model depends largely on the accurate modeling of the neural connectivity that mediates collinearity. Hence, accurate modeling of connections based on neuroscience, neurophysiology and psychophysics is of highest priority. 
In these studies, the extent and distance of the connections are measured with a parameter denoted as $\lambda$. The value $\lambda$ denotes a multiple of the wavelength of the Gabor filter. The background for this methodology is that the psychophysics community studies human subjects, and the subjects look at stimuli on a computer screen and the distances as well as sizes of these stimuli define lambda. We then estimate in our work here from the wavelength of the Gabor filter the size of the stimulus, as outlined in Appendix 2.1, because we also need to address other stimuli in our study. The estimation is based on several ways on the stimulus data of Maniglia et al. (2015) and Maniglia et al. (2015b) \cite{Maniglia2015a,Maniglia2015b}. As the result, we found that lambda can be justifully approximated by $2.0-2.4$ times the stimulus size (the precise value is ca. $2.1$ times). As an example, if a stimulus and thus Gabor kernel has a size of $11\times11$ pixels (the often-used value in our studies), $1 \,\lambda$ is $5$ pixels (we round all the numbers to full pixels to avoid discretization errors).

\begin{figure}[t]
    \centering
    \includegraphics[width=1.0\columnwidth]{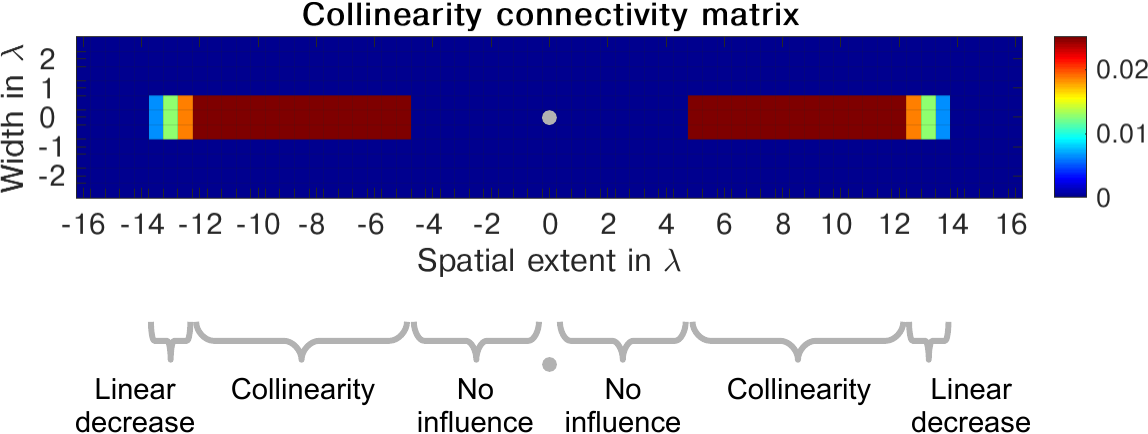}
    \caption{Precise spatial extent as well as width of the collinearity connectivity. The receiving neuron is marked in gray, and a red color denotes that a neuron at this position has a connection to the receiving neuron, i.e. the color denotes the weight strength. Both dimensions are measured in $\lambda$, with 1 $\lambda$ = size of wavelength of the Gabor, e.g. $1\,\lambda\,\hat{=}\,5\,$pixels. The connections in the direction of the spatial extent have a minimum distance, i.e. there is a region without a collinearity influence around the neuron, and the connections stretch until a maximum distance.  Note, only the connectivity matrix for a single orientation is shown, each orientation has its own connectivity pattern, which is rotated appropriately for each orientation. See main text for details (Sec. \ref{sec:modelNeuroAndKernel} and matrix $w_{x,x',l}$ in Eq. \ref{eq:col3}).}
    \label{fig:modelConnectMatrix}
    \vspace*{-0.25em}
\end{figure}

The collinearity influence is mediated by a certain extent of collinearity connectivity, and we found two kind of sources of biological data addressing our topic, Maniglia et al. (2015) \citep{Maniglia2015a}, and Kapadia, Westheimer \& Gilbert (2000) \citep{Kapadia2000}: The first source points to the range of the influence, which shows that collinearity in psychophysical studies has a maximum distance of influence, but also a minimum distance at which the influence first starts (Fig. \ref{fig:modelConnectMatrix}, spatial extent). Their data shows a start of the neural connections at 5$\,\lambda$ and an end at a maximum of 14$\,\lambda$. Given that a previous study \citep{Gilbert1998} found appropriate connections between the neurons, it is assumed that the neural connections reflect the psychological influence pattern. Note, in the literature, the exact start and the end of the connections/facilitation influence seems to vary slightly, for example, in the fovea (place of highest vision cell resolution) the relative distances are observed to be a slightly smaller (2$\,\lambda$ start, 12$\,\lambda$ end, \citep{Polat1993}) while Maniglia et al. measured in the close periphery 4$^\circ$ away from the fovea, and others measured also slightly deviating data (6-8$\,\lambda$ start, \citep{Maniglia2015a,Lev2011,Maniglia2011,Maniglia2015b}). We choose Maniglia et al. (2015) because for modeling purposes, it turns out in previous work to simulate as a compromise a slightly point away from the fovea \citep{Beuth2019,Antonelli2014}. Our models do not cover eccentricity-dependence and we like to cover all view field from the place of highest resolution (fovea) to the lowest resolution (far periphery), so it turns out to make a compromise and model the system after a resolution in between. In addition, we have to make some assumption as a data source for sake of simplicity of the model. 
Varying numbers could be easily incorporated in the model by changing the start and end of the connection matrix weights accordingly. 

Furthermore, the data of Maniglia et. al. show a decrease in influence at far extents, and we find that it can be modeled by a linear decrease in the weights from $10\,\lambda$ to $14\,\lambda$ (Fig. \ref{fig:modelConnectMatrix}a). In pixel space, given our example of $1\,\lambda\,\hat{=}\,5\,$ pixels, the minimum connection corresponds to 25 pixels and extends until 70 pixels. Note, the extent stretches in both directions.

With respect to the width of the connectivity pattern as well as its overall shape, we found the second source of work from Kapadia et al. (2000) \citep{Kapadia2000}, who studied the influence of a rotated, collinear line element in a paradigm called tilt illusion \citep{Solomon2004}. The experimental paradigm tests whether a slightly-rotated stimulus can be detected in the surround of a central stimulus. Kapadia et al. tested the spatial position of the flanker stimulus, giving us a 2D layout of the positions influencing the central element. We assume here that the tilt illusion is mediated by the same underlying connections and processing network as collinearity, and we use these data, since similar data are not available for the collinearity paradigm. As a result, the authors found a width of about 2$\, \lambda$, hence, we model the connectivity matrix with a width of 2$\, \lambda$ (Fig. \ref{fig:modelConnectMatrix}, width). In our beforehand-given example, this width corresponds to $10$ pixels. Furthermore, they systematically tested the position of the flankers, thus precisely determining the shape of the collinearity influence pattern: they found a rectangular influence region in the top view, i.e. the width of the influential area is at the closest position  the same width as at the far side. We model the connectivity matrix according to the data. Therefore, we could eliminate several free parameters of the model and we were able to constrain them with biological data. The design of the connectivity depends strongly the results of the evaluations and the model’s real-world suitability.

\begin{table}[t]
    \renewcommand{\arraystretch}{1.1}
    \centering
    \small
    \scalebox{0.85}{\begin{tabular}{|M{1.7cm}|M{1.7cm}|M{1.9cm}|M{1.0cm}|M{1.0cm}|}
        \hline
		Layer & Type & Output shape & Kernel size & Stride \\
        \hline
        \noalign{\vskip 2pt}

        \hline
        Collinearity layer & Collinearity & $513\times76\times9$ & / & / \\
        \cline{1-5}
        Pooling layer & Pooling & $513\times76\times9$ & $3 \times 3$ & $3$ \\
        \cline{1-5}
        Gabor Layer & Convolution & $1540\times229\times9$ & $11 \times 11$ & $1$ \\
        \cline{1-5}
        Image & Input & $1540\times229\times1$ & $ / $ & $ / $ \\
        \hline
    \end{tabular}}

    \caption{The collinearity model's sizes for the first real-world scenario (semiconductor wafer - fault detector). 
    Note, the different scenarios have different image resolutions, thus the model sizes differ.}
    \label{tab:collinearityModelNumbers}
    \vspace{-0.5em}
\end{table}

The psychological literature of collinearity shows that there exists also a suppressive influence in the close proximity of the target \citep{Maniglia2015a,Maniglia2015b,Polat1993,
Kapadia1995,Schmidt2009}. In other paradigms, these influences are also measured and denoted as surround suppression which has been widely observed in the visual cortex (e.g. \citep{Cavanaugh2002b, Sundberg2009}), and similarly, we have modeled in our earlier work such surround suppression influences \citep{Beuth2015a}. In our former modeling works, it turned out that it is beneficial to model different suppressive influences via different neural connections \citep{Beuth2015a}, even if they might be transferred over the same anatomical connection in the brain. As the surround suppressive influences are also found in other paradigms, we like to propose here to see surround suppression and collinearity as two different entities, despite suppression and   collinearity enhancement are often measured together.
Hence, we leave the surround suppression out here as we like to develop a model of collinearity, enhancing collinearity-aligned edges, for the computer vision community and we feel without surround suppression the model is more straight forward. We currently plan to include surround suppression via a separate connection in an upcoming neuroscience publication and plan it currently as future work.

Our proposed model is based regarding its structure, neurophysiology, and neuroanatomy on the primary visual cortex (V1) in the human brain \citep{Brodmann1909,Teichmann2018,Riesenhuber1999} (Fig. \ref{fig:modelConnections}, inset). The Gabor stage cells correspond to the cells in V1 reacting to edges \citep{Hubel1962}, whose visual receptive fields are approximately in the shape of a Gabor function, and are commonly simulated by Gabor filters \citep{Jones1987}. Note, these kind of Gabor filters are Gabors in the pixel space, and not in Fourier space, as often employed in the computer vision community. Concerning pooling, it is thought that a pooling aspect exists within the circuits of the visual cortex - in the primate brain, a cortical area consists of itself of up to 6 layers \citep{Brodmann1909}. It is thought that the cells in layers 2 and 3 aggregate spatially over the cells in layer 4 \citep{Beuth2015a, Douglas2004, Douglas2007}, as conducted by modeling the areas of the visual cortex \citep{Beuth2015a}. We also perform spatial pooling in our model in the pooling layer. Hence, in the model, we reproduce the classical structure of the primary visual cortex, an encoding of edges via Gabor filters in a deeper layer and a pooling of such responses in a higher layer.
The core component of the model is the collinearity connectivity. Studies found that the connections facilitating collinearity operate inside the same layer (lateral connections) \citep{Angelucci2002, Cass2005}, thus, we simulate the connections in a lateral style - side-by-side - inside our model’s collinearity layer. Other works found that connections in layers 2/3 of the primary visual cortex mediate collinearity \citep{Kapadia1995}, hence our collinearity layer might correspond to layers 2/3 of the visual cortex area. We opted for a separate entity for collinearity in the model since it allows a better control of the model, whereas in the cortex, both the model's collinearity layer and its earlier layer, i.e. the pooling layer, may correspond both to layers 2/3 of the primary visual cortex. A final assumption had to be made regarding the influence of the collinearity connections ($a\Area{Col}$ in Eq. \ref{eq:col1}, \ref{eq:col3}), we choose here to apply it via the term $x \cdot (1+a\Area{Col})$, a multiplicative upscaling. This term is the typical term by which visual attention in neuroscience influences the neural activity via feedback connections \citep{Beuth2019, Beuth2015a, Hamker2005b, Bayerl2007b}, and we thought in the human brain, collinearity might influence the signals in a similar way. 

The model can be downloaded at \url{https://github.com/fbeuth/collinearity/}.

\begin{figure}[t]
    \centering
    \vspace{-0.2cm}
    \hspace{0.1cm}
    \includegraphics[width=0.975\columnwidth]{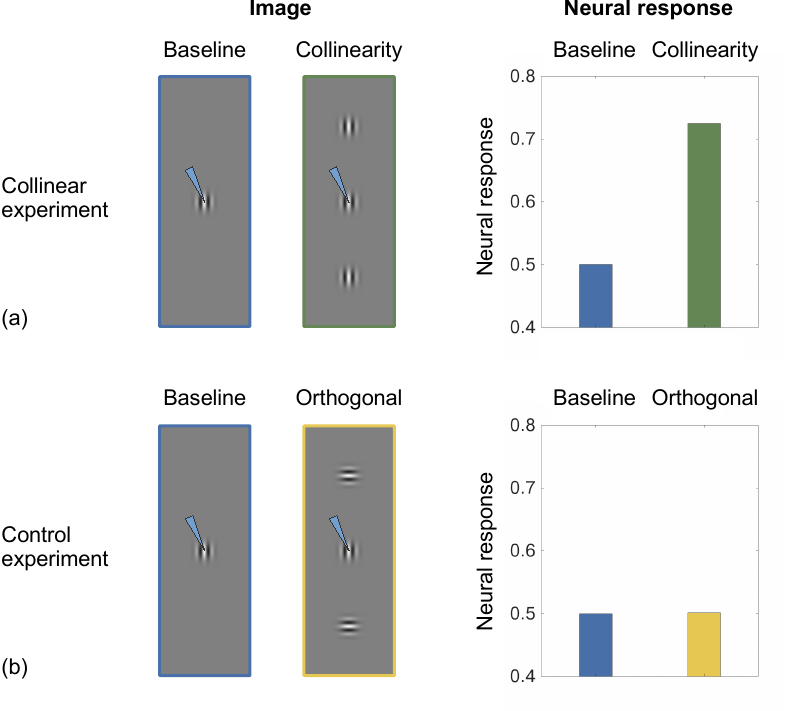}
    \caption{Basic example of operation of the collinearity principle. \textbf{(a)} If line elements are arranged along a longer line pattern, they are enhanced. The outer line elements are denoted as flankers. \textbf{(b)} If the outer flankers are not aligned, i.e. by $90^\circ$ rotated, they are not enhanced. On the left are shown the image setups, whereas on the right are shown the neural responses, which are read out for the central position of the collinearity layer (marked by the arrow, response $r\Area{Col}$, cf. Sec. \ref{sec:eq}).
    }       
    \label{fig:resultsBasic}
    \vspace*{-0.5em}
\end{figure}

\section{Results I}
\label{sec:results}

\subsection{Basic model behavior}
\label{sec:resultsBasic}

The basic example of collinearity is the enhancement of lines in the following setup of Fig. \ref{fig:resultsBasic}a. When a line is surrounded by similar-oriented neighbor lines in a linear fashion (collinear condition), the line is enhanced. 
Enhancement means that the central target element is better perceived, recognized, and processed. In the setup, the central line element is investigated (\textit{target}), while the outer line elements are called \textit{flankers}. The psychology literature (psychophysics, \citep{Maniglia2015a, Polat1993}) outlines this effect as a better perception of the central line element by humans. Psychological studies investigate this phenomenon with Gabor-shaped stimuli, which denotes the current state of the art, leading thus us to the replication of the Gabor-setup (Fig. \ref{fig:resultsBasic}a). The image in such a setup typically consists of a gray background. Here, we choose a gray value of 127.
The condition of collinearity is contrasted with a finding when the flankers are rotated by $90^\circ$, denoted control or orthogonal condition (Fig. \ref{fig:resultsBasic}b). This second experimental condition does not express an enhancement of the target's neural response.

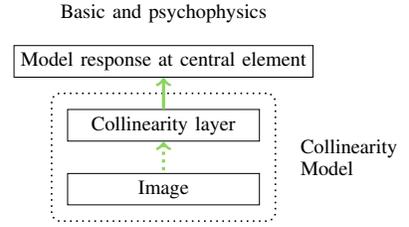
\begin{figure}[t]
	\centering
	\hspace*{1.4cm}
	\small
	\scalebox{0.85}{
    \begin{tikzpicture}[
        r/.style={draw, rectangle}]
    \node[r, minimum width=3cm, draw=none] (node5) at (1, -0.25) {Basic and psychophysics};
    \node[r, minimum width=3cm] (node4) at (1, -1) {Model response at central element};
    \node[r, minimum width=3cm] (node3) at (1, -2) {Collinearity layer};
    \node[r, minimum width=3cm] (node1) at (1, -3) {Image};
    \node[r, dotted, minimum width=3.5cm, minimum height=2cm, rounded corners, line width=0.25mm] (node0) at (1, -2.5) {};
    \node[r, draw=none, align=left] (node7) at (3.9, -2.5) {Collinearity \\ Model};
    \draw[<-, line width=0.5mm, arrowGreen] (node4) -- (node3);
    \draw[<-, dotted, line width=0.5mm, arrowGreen] (node3) -- (node1);    
	\end{tikzpicture}}
    \caption{Collinearity model in the use cases of the basic example, in the systematic evaluations, and in the psychophysics. The layers inside the collinearity model are not illustrated and its internal processing is symbolized by the dotted arrow (cf. model figure \ref{fig:model}).}
    \label{fig:modelcaseBasic}
    \vspace*{-1.5em}    
\end{figure}

In our proposed model, the enhancement is mediated by connections at the level of the collinearity layer, which connects neurons with the same feature that are arranged in a line (see Sec. \ref{sec:model} for details). The connections multiplicatively enhance the target's neural response when the neurons at the flanker positions are active. The connectivity is designed within the stage collinearity (denoted lateral between neurons) since it is though that the connectivity is mediated by lateral connections at the level of complex cells in the primary visual cortex in the human brain \citep{Kapadia1995}, hence reflecting the known connectivity aspects in the brain.

Methodologically, a single value from the collinearity layer is read out in this basic setup. The central element in the image is for interest in this task, whereby there is typically a single orientation present in such an image setup, thus we read out the value of the neuron in the collinearity layer $r_{x_1,\,x_2,\,l}$ at the central position ($x_1=N/2,\,x_2=M/2$) and from the only presented orientation ($l$). This approach is used in the first experiment and all systematic evaluation experiments in the following section (Fig. \ref{fig:modelcaseBasic}). Alternatively, we can read out the responses along a slice through the image, for example at an edge, to better understand the principle of collinearity. This idea is performed in the systematic evaluation experiments too.

\subsection{Systematic evaluation of the model}
\label{sec:resultsSystematic}

In the first block of analyses, we systematically evaluate the collinearity method to better understand the principle and to measure its behavior. Exemplary, we choose analyses regarding the influence of (i) the image contrast, (ii) the rotation and the alignment of the collinear structures, and (iii) the length of the line-like structures.

\subsubsection{Contrast influence I}
\label{sec:resultsContrast}

\begin{figure}[t]
    \centering
    \includegraphics[width=1.025\columnwidth]{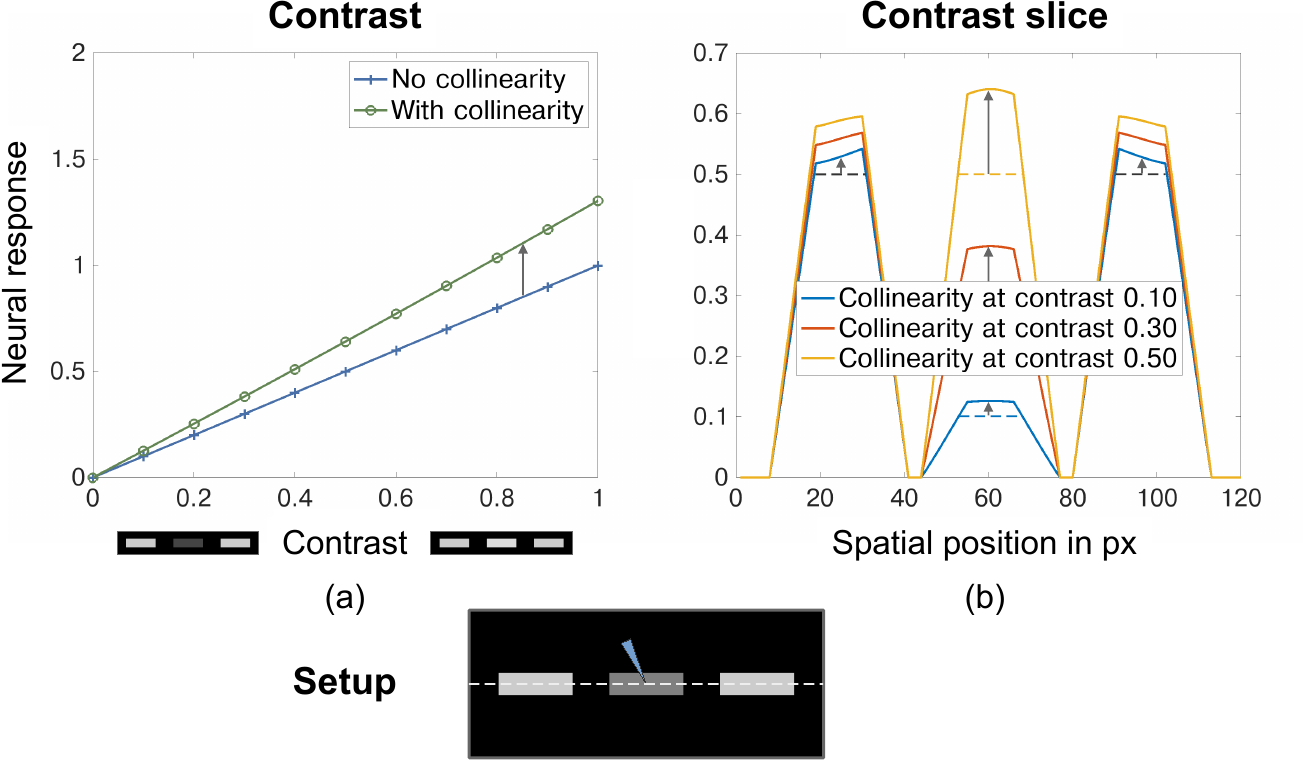}
    \caption{Evaluation of the contrast's influence on the collinearity approach via a low-visible line element. 
    \textbf{a)} Neural response of the system depending on the contrast of a central line element between two outer lines (flankers). The central line element is simulated with varying illumination in front of a black background, and the setup is shown as an inset. \textbf{b)} Neuronal response along a slice on the line (shown as a dotted line in the inset), illustrated at different contrasts of the middle line segment. The dotted lines in (b) illustrate the neural responses without collinearity. The middle part denotes the target element, whereas the left and right denote the flankers.} 
    \label{fig:resultsContrastGap}
    \vspace*{-0.5em}
\end{figure}

First, our collinearity approach is analyzed regarding the contrast. Since the principle of collinearity enhances the detection of stimuli, it is of interest to investigate the detection of low-contrast line elements or whole lines by flankers and collinearity. In our first analysis, we examine the detection of a low-contrasted line element between two outer, well-visible flankers (surrounding lines, Fig. \ref{fig:resultsContrastGap}). The central line segment has been simulated with systematically varied illuminations in front of a black background, while the flankers have a fixed illuminance of $127$. The contrast of a stimulus in the image ($0 ... 1$) is calculated as usually: $C = \left(|I_{stimulus} - I_{background}|\right)/255$, where $I_{stimulus}$ is the intensity of the stimulus and $I_{background}$ is the intensity of the background. The increasing contrast triggers a rise in the signal strength towards the model's first layer's Gabor neurons. The collinearity experiment is then compared to a control experiment in which we process the central line segment without collinearity.  In both conditions, we read out a single value from the collinearity layer at the position of the central element (Fig. \ref{fig:resultsContrastGap}a), or along a slice through the image space (Fig. \ref{fig:resultsContrastGap}b), as described beforehand in Sec. \ref{sec:resultsBasic}.
 
In our analysis, we observe an enhancement of the response through collinearity at all input contrasts  (Fig. \ref{fig:resultsContrastGap}a). Apart from that, we found a standard enhancement - with increased input contrast, it leads to an increase in collinearity neural activity, whereby it enhances multiplicatively. The flankers provide a constant source of influence towards the middle line segment, as they have a constant contrast level, hence, the enhancement has across all input contrasts of the central line the same multiplicative level.
Additionally, we present a spatial left-to-right line profile of the neural activity at differently selected contrast levels represented by different line colors in Fig. \ref{fig:resultsContrastGap}b. In this figure, the central element can be seen at different contrast levels as well as the flankers at constant contrasts. Note that slight border effects are visible at the flankers. All the slice analyses are performed at the level of the collinearity layer.

\begin{figure}[t]
    \centering
    \includegraphics[width=1.025\columnwidth]{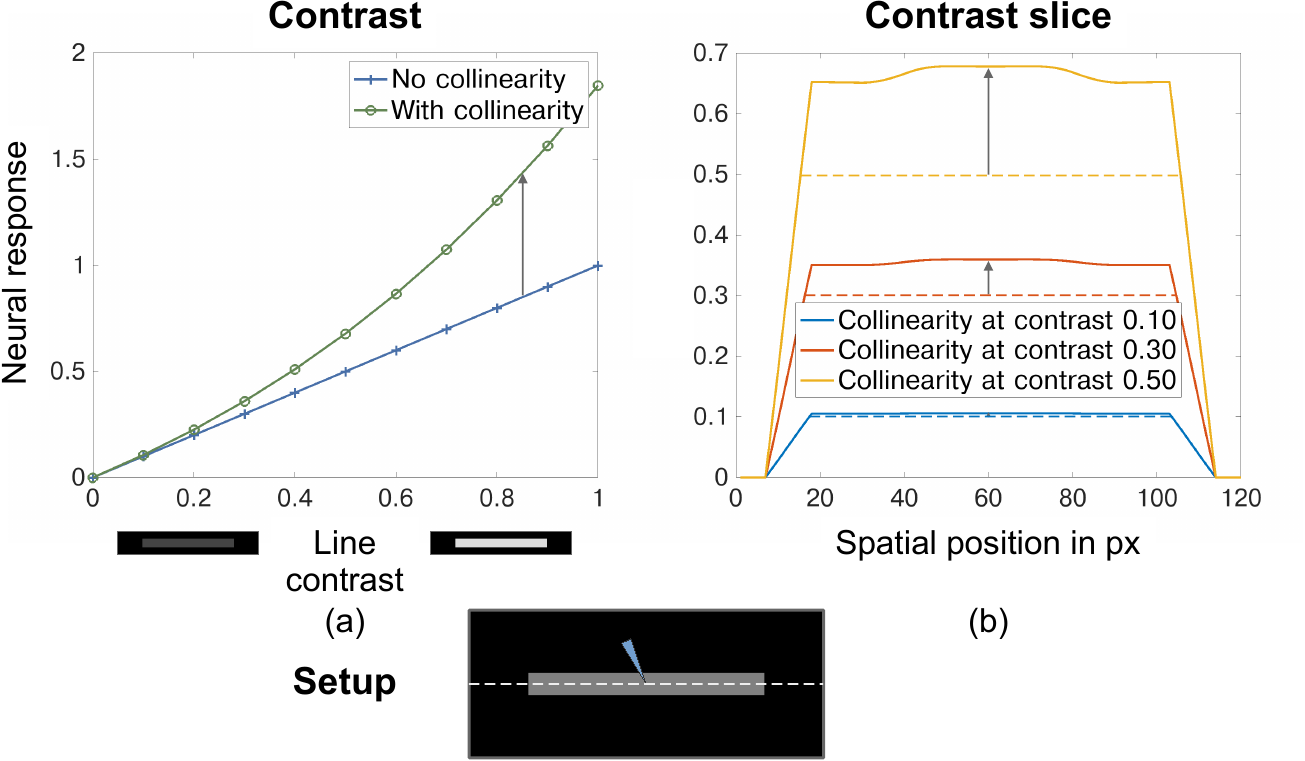}
    \caption{Influence of the contrast on the collinearity approach via a full line. \textbf{a)} Neural response of the system depending on the input's contrast. The input stimulus is an entire line with the given illumination in front of a black background as shown by the setup inset. \textbf{b)} Neuronal response along a slice on the line, shown at different contrasts of the input line. The notation of the figure is similar to Fig. \ref{fig:resultsContrastGap}.}
    \label{fig:resultsContrast}
    \vspace*{-0.75em}
\end{figure}

\subsubsection{Contrast influence II}

As a second experiment, we present a full line at a low contrast, and we repeat the previous experiment (Fig. \ref{fig:resultsContrast}a). The outer parts of the full line (i.e. the left and right parts of the line) serve as flankers and enhance the central element of the line. In our analysis, we observe an enhancement at all of the low contrasts, while we also observe a nonlinear enhancement, compared to the control condition. The reason for this is that the entire line is amplified, and the left and right parts of the line serve as flankers and these flankers are also amplified, which in turn leads to amplification of the central line parts. The effect is therefore non-linear.
In addition, we present a slice through the line - from left to right - with different selected contrast levels too, showing the neuronal activity (Fig. \ref{fig:resultsContrast}b). From this, it can be seen that the centered part of the line receives a basal contrast enhancement (e.g. a $0.5$ contrast is increased to a response of ca. $0.65$). The ends of the line have slight border effects. The border effects stem from the fact that parts of collinearity connectivity cover areas beyond the end of the line, and thus, the connectivity does not transmit any signals due to a lack of stimuli. 

Therefore, the collinearity principle is able to enhance both low-contrast line gaps as well as low-contrast lines.

\begin{figure}[t]
    \centering 
    \includegraphics[width=1.025\columnwidth]{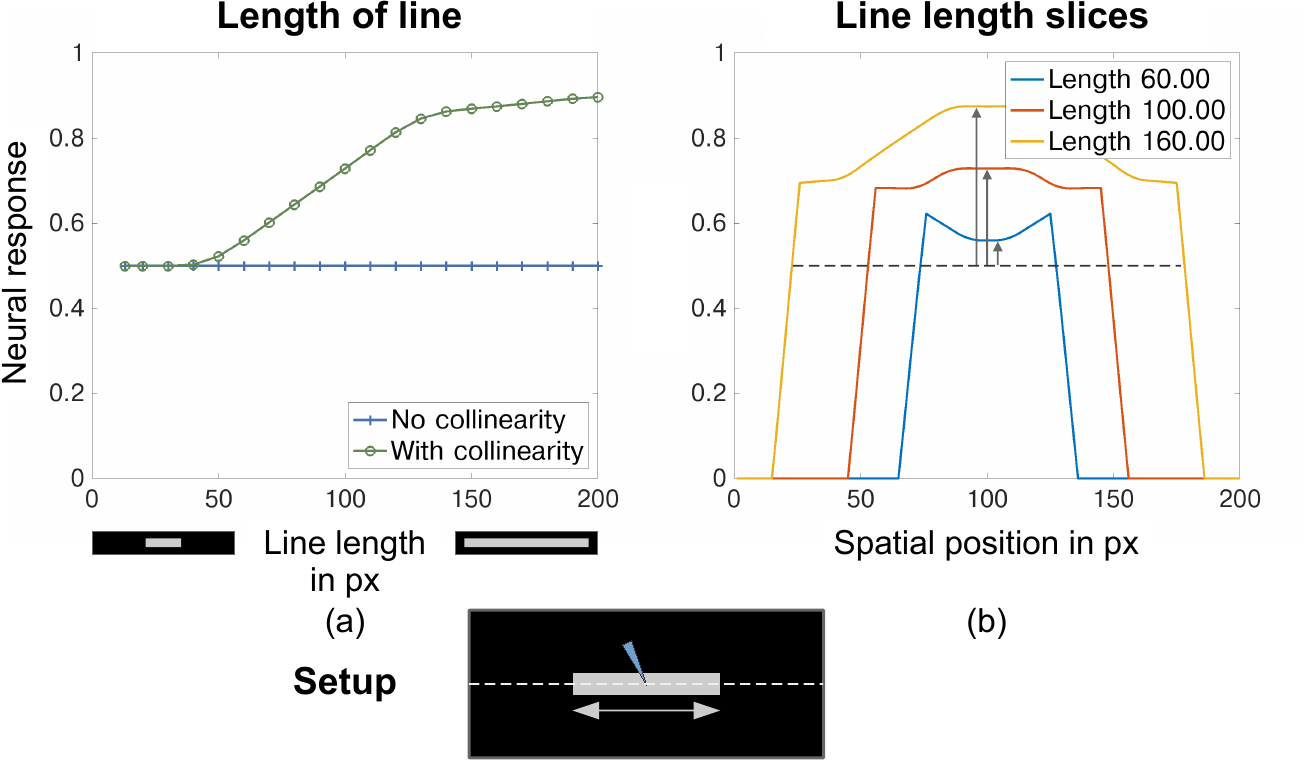}    
    \caption{Influence of the line's length on the collinearity characteristics. \textbf{a)} Collinearity response dependent on the line-segment's length in pixel. The stimulus is a line of the given length at a contrast of C=$0.5$ in front of a black background, as shown by the setup inset. \textbf{b)} Collinearity response along a slice on the line for three exemplary lines with the denoted lengths. The input contrast is always C=$0.5$.
    The collinearity connectivity has a minimum and maximum connection length, which influences the response. The connectivity is modeled based on the psychology (Sec. \ref{sec:modelNeuroAndKernel}).}
    \vspace*{-0.75em}
    \label{fig:resultsLength}
\end{figure}

\subsubsection{Influence of the line length}
\label{sec:resultsLength}

Next, we systematically investigate the collinearity method regarding the influence of the elongation of the collinear line. It will also answer such questions which are optimal lines for collinearity, what is the maximum distance, and what is the general behavior of the collinearity principle. We systematically vary the elongation of a line and measure the neural response in the collinearity layer (Fig. \ref{fig:resultsLength}). Furthermore, it is also interesting as the underlying collinearity connectivity has a minimum and maximum connection length, which influences the neural response, due to the fact that our connectivity is modeled based on psychology.
Methodologically, we read out a single value from the collinearity layer at the position of the central element (Fig. \ref{fig:resultsLength}a), or along a slice through space (Fig. \ref{fig:resultsLength}b), as described beforehand in Sec. \ref{sec:resultsBasic}. 
We found the following outcome(s) (Fig. \ref{fig:resultsLength}a): The influence of collinearity increases constantly with line length as more ‘line’ parts are aligned collinearly. In longer lines, the enhancement saturates to an upper limit as the underlying collinearity connectivity reaches its maximum range. In our test case, it reaches the maximum at a range of approximately 140 pixels. This corresponds with the theoretical expectation, as the collinearity connections have a maximum range of $14\, \lambda$ in both directions, and this value is: $14 \, \lambda \cdot 2 \cdot 5 \,\frac{pixels}{\lambda} = 140$ pixels. We observe interestingly that the response still climbs weakly after ca. 140 pixels. We found that this was caused by the ends of the line also receiving collinearity enhancements, which in turn enhance the central parts of the line more. The ends' enhancement only saturates at longer lines, hence, as a result, the entire structure also receives beyond the end of the collinearity's connectivity an enhancement. In addition, we again show a slice along the line - from left to right - of the neural activities (Fig. \ref{fig:resultsLength}b), selected for three exemplary line lengths.

\subsubsection{Replication of psychophysical data}
\label{sec:psychophysics}

\begin{figure}[t]
    \centering
    \includegraphics[width=1.04\columnwidth]{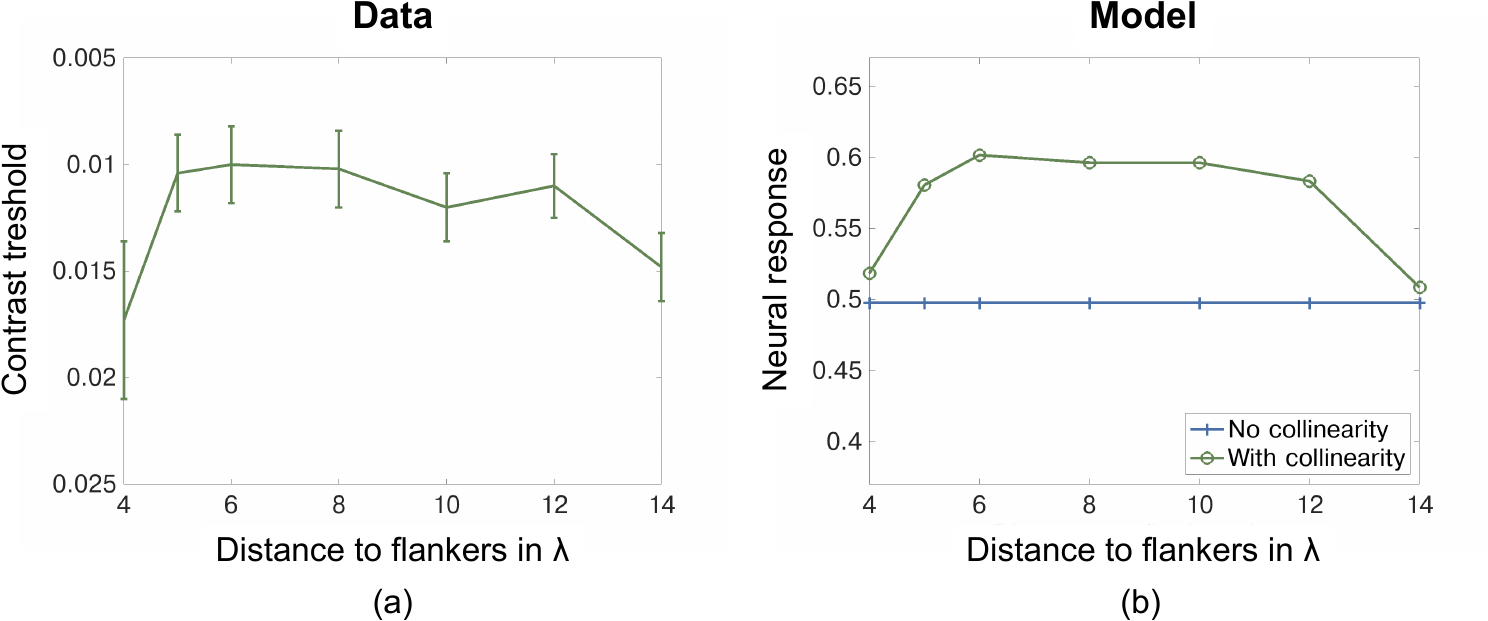}
    \caption{Replication of psychophysical data. Dependency of the collinearity effect on the distance between the target stimulus and its neighbors (denoted as flankers). \textbf{a)} Psychophysical data of the collinearity facilitation, from \citep[Fig. 3a]{Maniglia2015a}. Note, the illustration is plotted with an inverted y-axis as there was conducted a contrast-threshold experiment in the original study, in which a lower contrast threshold denotes a better detection of the stimulus. \textbf{b)} Influence of the collinearity facilitation in the model. The response is compared to the Gabor pooling response, which denotes an unmodulated baseline. }
    \vspace*{-0.5em}
\label{fig:resultsPsychophysics1}
\end{figure}

We replicate human behavioral data (psychophysical data) at the next step. The work of Maniglia et al. (2015) \citep{Maniglia2015a} points out that the collinearity effect has a minimum and maximum spatial range (other works point this out too). Their data shows that the facilitation by collinearity starts at 5$\, \lambda$ and ends at 14$\, \lambda$ (Fig. \ref{fig:resultsPsychophysics1}a). In their experiment, the stimulus was a Gabor patch (denoted target) with two neighboring stimuli at a variable distance (flankers). The distance between the target and the flanker was systematically varied. The separation was measured as a multiple of $\lambda$, denoting the wavelength of the Gabor patch. The study reported the contrast threshold at which human subjects can detect the target Gabor patch (contrast-threshold experiment). With collinearity, the subjects can detect the stimulus better, i.e. at lower contrasts. We replicated the data with our model (Fig. \ref{fig:resultsPsychophysics1}b). Methodologically, we use the same setup as in the original study by creating an input image with Gabor patches, and process the image with our system. In our model, we are able to directly record the neural activities at the target position instead of the perceived contrast of human subjects, and we read out the neural activity at the target position. Note, there has been also reported suppression for shorter relative distances of $\lambda$ ($\le4\,\lambda$), an effect which is left out in the model since we think suppression and collinearity facilitation might be two different entities (see Sec. \ref{sec:modelNeuroAndKernel} for details).
The reported effects of a maximum and minimum range of collinear facilitation have also been reported by other studies, e.g. \citep{Polat1993}. Note, this replication of the data allows us to develop the connectivity range of the collinearity connections in our model precisely according to the psychophysical data.

\subsubsection{Influence of the matching rotation of the flankers}
\label{sec:resultsRotation}

\begin{figure}[t]
    \centering 
    \includegraphics[width=0.9\columnwidth]{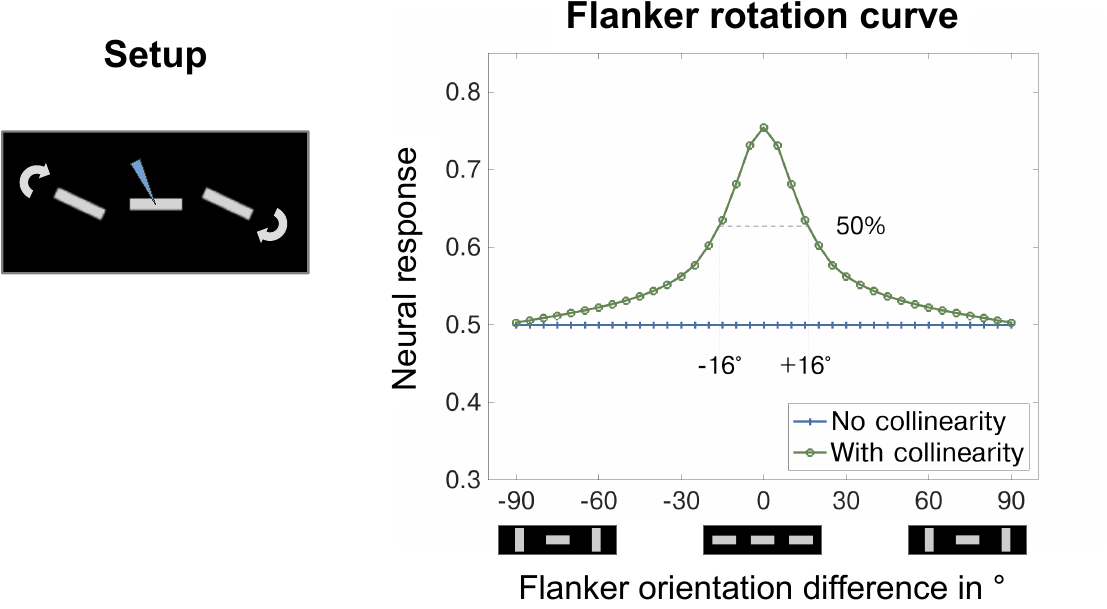}
    \caption{Influence when the flankers do not match precisely in the orientation (rotation tuning curve). The setup is a central stimulus with fixed orientation and accompanied by two flanker stimuli with systematically-changed orientation. The gray dashed line shows a 50\% value of the maximum.}
    \label{fig:resultsRotationTuning}
\end{figure}

In many application cases, the neighboring stimuli (flankers) will not match precisely the orientation of the target (Fig. \ref{fig:resultsRotationTuning}). To investigate what happens when the flankers do not match precisely, we obtain a tuning curve of the neural activity when the flanker stimuli change their orientation. In our test setup, we fix the orientation of the central, target element, and we rotate the flankers systematically in $5^{\circ}$ steps. We observe the following curve as illustrated in Fig. \ref{fig:resultsRotationTuning}. The curve indicates that it is up to ca. $\pm 15^{\circ}$ until the neural response drops below $50\,\%$ of its maximum value, showing that some invariance exists regarding the flankers' orientations. This is an essential finding and investigation for the case that the flanker does not match the middle line segment precisely since it often occurs in real-world data.

--- Structure dependent enhancement of contrasts ---
Concluding, the collinearity principle enhances edges and other image elements when they are arranged in a larger spatial super structure. This can include edges arranged on a longer line (Sec. \ref{sec:resultsBasic}, \ref{sec:resultsContrast} - \ref{sec:resultsRotation}), or simply longer edges (Sec. \ref{sec:resultsLength}). When the edges are not aligned, i.e. to some degree rotated/not-aligned (this section) or simply orthogonal-rotated (Sec. \ref{sec:resultsBasic}), the image element is not enhanced. Due to these facts, we see the collinearity principle as a structure-dependent enhancement of the image. Contrary to normal computer vision approaches, in the simplest case contrast normalization via histograms \citep{Szeliski2022}, collinearity enhances the image structures depending on the neighborhood, and hence, it puts the context of the image into help.

\section{Results II: Real-world evaluations}
\label{sec:resultsRealworld}

In the next sections, we test the collinearity principle in various real-world computer vision applications to explore its utilization. In parallel, we sketch the deployment of the collinearity model within several methodologies, choosen as deep learning (two times), saliency models, and classical computer vision.

\subsection{Use case of fault detection in semiconductor wafers}
\label{sec:resultsWafer}

\subsubsection{Domain overview and feature detector}

\begin{figure}[t]
    \centering  
    \input{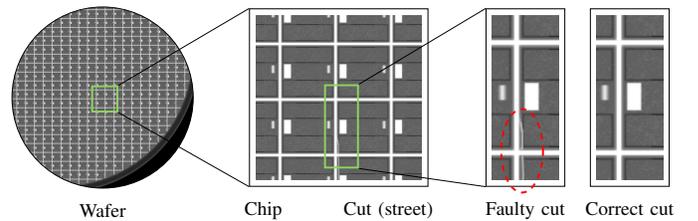}
    \vspace*{-0.55cm}
    \caption{Semiconductor wafer domain. Overview example of a wafer (left) with chips (middle) and faulty and correct cuts (right). The image classification task is to distinguish faulty cuts from correct cuts (the fault is marked in red). The examples shown were generated synthetically from the original images to protect the intellectual property. Reprinted and adapted from \cite{Beuth2021}.}
    \label{fig:resultsWaferExample}
\end{figure}

As the first use case, we investigate the collinearity principle's usage in the field of fault detection in semiconductor wafers. 
In this scenario, a silicon wafer is cut by laser- or sawing-technology to be separated into single chips  (Fig. \ref{fig:resultsWaferExample}) \citep{Huang2015a, Beuth2021, Beuth2020a}. During the cutting process, the cut can leave the prescribed cut line, called street, and travels incorrectly through the chip, denoting problems in the production process. Recognizing such faults early on in the production process saves costs, reduces the workforce, and increases the yield of chips from a wafer \citep{Huang2015a}. Figure \ref{fig:resultsWaferExample} illustrates a correct and a faulty cut. We analyze this application case since the correct cuts are collinear structures, which inspires us to explore the collinearity principle based on the case. 

\begin{figure*}[t]
    \centering  
    \includegraphics[width=0.95\textwidth]{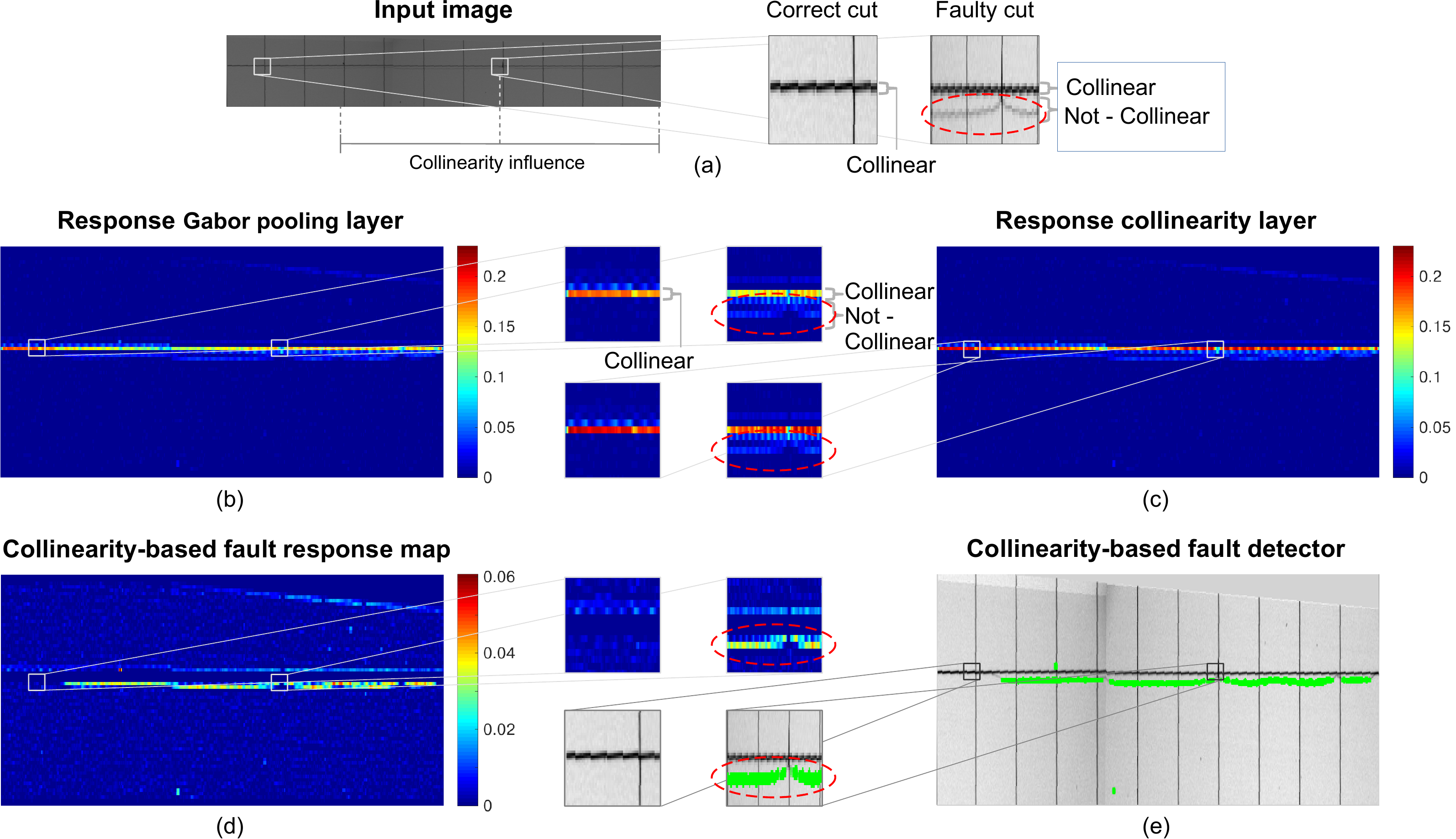}
    \caption{Fault detection in the semiconductor wafer use case via a feature detector. \textbf{a)} Example image. The correct cut lines of the wafer are collinear, whereas the faulty cuts (marked in red) are non-collinear. The goal is to distinguish correct from faulty cuts. 
    \textbf{b)} Response of the Gabor layer (in its pooled version). \textbf{c)} Response of the collinearity layer. For each response map, the values are illustrated as the maximum over all orientations. 
    \textbf{d)} Developed collinearity-based fault detection solution. A fault response map is calculated from the difference between the collinearity layer response (c) and the Gabor layer response (b).
    \textbf{e)} With appropriate thresholding, a fault detector map is created from the fault response map (d). Note, the image is highlighted for better visibility via contrast normalization, and the collinearity model receives the unpreprocessed image (a). All plots in (b)--(e) are stretched in y-direction for higher visibility.} 
    \label{fig:resultsWafer}
    \vspace*{-0.5em}
\end{figure*}

\begin{figure}[t]
	\centering
    \scalebox{0.85}{
    \small
    \begin{tikzpicture}[
        r/.style={draw, rectangle}]
    \node[r, minimum width=3cm, draw=none] (node6) at (1, 0.75) {Fault detection in semiconductor wafers};
    \node[r, minimum width=3cm,text depth=.2em] (node5) at (1, 0) {Fault detector};
    \node[r, minimum width=3cm,text depth=.2em] (node4) at (1, -0.75) {Fault response map};
    \node[r, minimum width=3cm,text depth=.2em] (node3) at (1,  -1.5) {Difference map};
    \node[r, minimum width=3cm,text depth=.2em] (node2) at (1, -2.45) {Collinearity layer};
    \node[r, minimum width=3cm,text depth=.2em] (node1) at (1, -3.20) {Image};
    \node[r, dotted, minimum width=3.5cm, minimum height=1.5cm, rounded corners, line width=0.25mm] (node0) at (1, -2.825) {};
    \node[r, draw=none, align=left] (node7) at (3.9, -2.825) {Collinearity \\ Model};
    \draw[<-, line width=0.5mm, arrowGreen] (node3) -- (node2);
    \draw[<-, line width=0.5mm, arrowGreen] (node4) -- (node3);
    \draw[<-, line width=0.5mm, arrowGreen] (node5) -- (node4);    
    \draw[<-, dotted, line width=0.5mm, arrowGreen] (node2) -- (node1);
	\end{tikzpicture}}
    
    \caption{Collinearity model in the use case of fault detection in semiconductor wafers - feature detector. The neural layers inside the collinearity model are not illustrated and its internal processing is symbolized by the dotted arrow.}
    \label{fig:modelcaseFeaturedetector}
\end{figure}
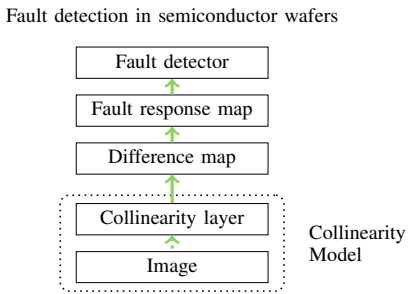

Based on this knowledge, we show the development of a collinearity-based detector system, and its work-flow is illustrated in Fig. \ref{fig:modelcaseFeaturedetector}. 
On an exemplary sample image, the collinearity processing is shown in Fig. \ref{fig:resultsWafer}. The model processes the image with the Gabor layer (Fig. \ref{fig:resultsWafer}b), pools the response, and then applies the collinearity processing (Fig. \ref{fig:resultsWafer}c). As illustrated by Fig. \ref{fig:resultsWafer}, the streets (intended cut lines) are collinear, whereas the faults are not collinear (the faults are located below the intended cut line in the figure, marked in red). Hence, the differing property between the classes is that the correct cuts are collinear structures and the faulty cuts are not-collinear.
The fault detector is developed to react to everything that is \textit{not-collinear} (Fig. \ref{fig:resultsWafer}d). It is implemented by taking the difference between $r\Area{Col}$ (collinear activity) and $r\Area{Pooling}$ (pooled Gabor activity), as a measurement of the collinearity enhancement (difference map). If this enhancement by collinearity remains below a certain threshold, then the detector reacts as an indication of non-collinearity.
We choose this development way since non-collinearity defines the faults, which shall be recognized, in this application case. By applying an appropriate threshold, a fault detector is created from the fault response map, and it outputs a map of errors in the silicon wafer (Fig. \ref{fig:resultsWafer}e).

We found that the detector is able to recognize the cuts pretty well in our example image (Fig. \ref{fig:resultsWafer}e). In the faulty cuts, the three large ``arcs'' of the faulty cuts are all detected correctly, and the small arc on the right of the image too. The main idea is to exploit that a differentiating property of class correct and class faulty cuts is that fine cuts are collinear and faulty cuts are not (or at least to a much lower degree). A part from that, the detector reacts for example also to a small hole at a street crossing, which is indeed turned out as a fault (breakage) after deeper analysis. The sketched approach has the advantage that it is computationally very low-cost, making it ideal for applications within the production process or in low-energy devices. Yet, it is also not perfectly robust alone, which is visible due to the noise in the detector's response map. In summary, we conclude that the principle of collinearity can indeed be successfully applied to this case, and we sketch here the deployment as a detector and show qualitative results in the use case of defect detection in semiconductor wafers.

\subsubsection{Deep learning}
Secondly, we explore the combination of collinearity with deep learning within the use case (deep learning, \citep{LeCun2015}).
Furthermore, the idea is motivated due to the imperfect robustness of the detector as a standalone, and we would like to measure the collinearity power also quantitatively. We follow the methodology of Beuth et al. (2021) \citep{Beuth2021}, who have recently applied deep neural networks to the defect detection of semiconductor wafers and since they constitute one of the current state-of-the-arts \citep{Beuth2021, Schlosser2022}. One crucial idea of the original authors has been that to put the cropped street images, which have a higher resolution (Fig. \ref{fig:resultsWaferExample}) into the deep neural network, and not the full chip images. 
Other approaches based on deep neural networks can be found for example in \citep{Cheon2019}. 

\begin{figure}[t]
	\centering
	\scalebox{0.85}{
    \small
    \hspace{0.5cm}
    \begin{tikzpicture}[
        r/.style={draw, rectangle}]
    \node[r, minimum width=3cm, draw=none,text depth=.2em] (node5) at (2.6, 0.75) {Deep learning in semiconductor wafers};
    \node[r, minimum width=3cm,text depth=.2em] (node4) at (2.6, 0) {Class};
    \node[r, minimum width=3cm, minimum height=0.87cm,text depth=.2em] (node3) at (2.6, -0.95) {Deep neural network};
    \node[r, minimum width=2.65cm,text depth=.2em] (node2) at (1, -2.45) {Image};
    \node[r, minimum width=2.65cm, align=center,text depth=.2em] (node1) at (4.2, -2.45) {Feature detector \\ (Fault response map)};
    \node[r, draw=none, align=left] (node6) at (0.6, -1.75) {Channel 1};
    \node[r, draw=none, align=left] (node7) at (4.6, -1.75) {Channel 2};
    \node[r, minimum width=2.65cm,text depth=.2em] (node8) at (4.2, -3.4) {Difference map};
    \node[r, minimum width=2.65cm,text depth=.2em] (node9) at (4.2, -4.35) {Collinearity layer};
    \node[r, minimum width=2.65cm,text depth=.2em] (node10) at (4.2, -5.1) {Image};
    \node[r, dotted, minimum width=3.15cm, minimum height=1.5cm, rounded corners, line width=0.25mm] (node0) at (4.2, -4.725) {};
    \node[r, draw=none, align=left,text depth=.2em] (node11) at (6.85, -4.725) {Collinearity \\ Model};
    \draw[<-, line width=0.5mm, arrowGreen] (node3) -- (node1);
    \draw[<-, line width=0.5mm, arrowGreen] (node3) -- (node2);
    \draw[<-, line width=0.5mm, arrowGreen] (node4) -- (node3);
    \draw[<-, line width=0.5mm, arrowGreen] (node1) -- (node8);
    \draw[<-, line width=0.5mm, arrowGreen] (node8) -- (node9);
    \draw[<-, dotted, line width=0.5mm, arrowGreen] (node9) -- (node10);
	\end{tikzpicture}}    
    \caption{Collinearity model in the use case of deep learning in semiconductor wafers (configuration 1). The fault response map is based on the modeling work shown in Fig. \ref{fig:resultsWafer}d, and the notation of the figure is similar to Fig. \ref{fig:modelcaseFeaturedetector}.}
    \vspace*{-0.25em}
    \label{fig:modelcaseDLconf2}
\end{figure}
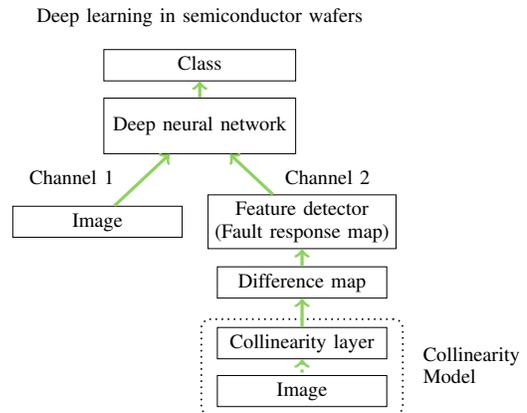

We test two experimental approaches to apply collinearity to deep neural networks. 
In the first approach, we replace the first stage of the deep neural network with a combination of the image and a collinearity feature-detector, i.e. the collinearity-based fault response map (work-flow shown in Fig. \ref{fig:modelcaseDLconf2}). The idea was inspired by findings that the first stage of a deep neural network can be replaced with Gabor filters extracting local features \citep{Evans2022,Dapello2020}. Hence, the collinearity serves as a secondary information source for the DNN.

The model configuration is given by the equation: 
\begin{eqnarray}
   I\Area{Out}_{x_1,\,x_2,\,2} &=& \mbox{channelwise concatenation of} \left\{I_{in}, r\Area{E} \right\}
\end{eqnarray}
Where $I_{in}$ is the original gray input image from the data set, and $r\Area{E}$ is the response of the fault response map (cf. Fig. \ref{fig:resultsWafer}d), respectively. Note, the map $r\Area{E}$ needs first to be resized to image resolution via interpolation as the map's resolution is smaller due to the pooling operation. 

In the second alternative approach, we test to suppress the irrelevant image-features and -parts for the application task (work-flow in Appendix Fig. 10). The ``streets'' structures are irrelevant for the current application case and they have collinearity properties, hence, their high level of collinearity is leveraged to suppresses undesired information. We developed a feature extractor that reacts to everything that is collinear, and it then blurs out these image parts by setting selected proportionally pixels to a gray value. The modeling approach is given by the following Eq. \ref{eq:waferConf1a} -  \ref{eq:waferConf1d}. 
In them, $D$ denotes the difference between the collinearity response and its input, the Gabor response, and it thus encodes the enhancement through collinearity. $I_{in}$  denotes the original gray-scale image, $r\Area{Gabor}$ the Gabor layer response, $r\Area{Col}$ the collinear layer response, $\odot$ elementwise multiplication, and resizing is also required. $I_{mean}$ denotes the gray value calculated as the average intensity of the image. 
The Eq. \ref{eq:waferConf1d} sets pixel values with a high amount of collinearity, which is given by $D$, to gray, and it keeps pixel values with a low amount of collinearity. 
\begin{eqnarray}
    D_{x_1,\,x_2} &=& \max_{l'} \left( r\Area{Col}_{x_1,\,x_2,\,l'} - r\Area{Gabor}_{x_1,\,x_2,\,l'} \right)  \label{eq:waferConf1a} \\
    && \mbox{Normalize D to $[0,1]$} \\
    D'_{x_1,\,x_2} &=& \left\{\begin{array}{ccc}D & &  D \ge 0.2 \\ 0 &&  D < 0.2 \end{array} \right.\\
    I\Area{out}_{x_1,\,x_2} &=& I_{mean} \cdot D' + I_{in} \circ (1-D') \label{eq:waferConf1d} 
\end{eqnarray}

The CNN classifies images into classes faulty and flawless, and it is custom made, on the basis of the VGG network \citep{Simonyan2015} (see Appendix Tab. 1). 
To exploit the long-range connectivity of the collinearity processing principle, it is important to note that the collinearity model runs across the whole wafer image that has a resolution of about $10000 \times 10000$ pixels. 
As experimental details, we split the dataset randomly into $50\,\%$ training, $25\,\%$ validation, and $25\,\%$ test set, employ a batch size of 48, and utilize the Adam optimizer with a learning rate of $0.01$. The total number of test samples is $ 3\,217$, distributed over both classes \{$C_0, C_1$\}. 
As ablation study, we re-conduct our experiments without collinearity from our previous study \citep{Beuth2021,Beuth2020a}. It utilizes the same CNN as in our collinearity experiments, yet without the collinearity information (input channel 2), and uses the same data set.

We found the following outcomes of the performance (Tab. \ref{tab:resultsWaferAcc}): Approach 1 works well, whereas approach 2 does not result in an improvement of the classification performance. The number of incorrectly detected images decreases from 209 to 169 items and the accuracy increases from $93.50\,\%$ to $94.74\,\%$ respectively. The error rate thus decreases from $6.50\,\%$ to $5.26\,\%$, which denotes an improvement factor of 1.24x. Note, in a post-hoc analysis, we observed that only up to 50\% of the faults in the data set are such depicted lines that potentially benefit from collinearity. Thus, we observed that the amount of improvement depends on the number of collinear-shaped structures in the data set, and real-world data sets might often have a mixture of faults.

\begin{table}[t]
    \caption{Results of the fault detection in semiconductor wafers via deep learning. Two configurations were tested (Configurations 1 and 2), while the baseline refers to IECON - Beuth et al. (2021, 2020) \citep{Beuth2021,Beuth2020a}. All results were measured and averaged over 5 runs.}
    \label{tab:resultsWaferAcc}
    \centering
    \begin{tabularx}{\columnwidth}{ L{4.5cm} P{1.7cm} P{1.7cm}}
        \toprule
        Model & Acc. \{$C_0, C_1$\} & Error rate \\
        \midrule
        CNN without collinearity (IECON 2020 \citep{Beuth2021,Beuth2020a}) &        $93.50\,\% \newline \pm 1.46~~$ & $6.50\,\% \newline \pm 1.46~~$  \\
        \midrule
        CNN with collinearity (Ours) \newline Configuration 1 & $\textbf{94.74\,\%} \newline \pm \textbf{1.09}~~$ & $\textbf{5.26\,\%} \newline \pm \textbf{1.09}~~$ \\
        \midrule
        CNN with collinearity (Ours) \newline Configuration 2 & $93.04\,\%         $ & $6.96\,\%$           \\
        \bottomrule
    \end{tabularx}
    \vspace*{-0.25em}   
\end{table}

In conclusion, we found that collinearity operate well with deep learning, it also improves quantitatively the performance in the use case of semiconductor wafers. We illustrated under which circumstances this neuro-science principle is useful, and as a use case, we shown the deployment of the principle in combination with deep neural networks.

\textit{Discussion:} The core parameters, which might be adapted by a process engineer for the task at hand, are the parameters of orientation and size of the kernel. In the application case, since the correct cuts are oriented horizontally, and faults deviate in orientation to some degree from them, we found that suitable results are achieved by choosing the orientations $\theta = \{-20,\,-15,\,-10,\,-5,\,0,\,5,\,10,\,15,\,20^\circ \}$. The orientations are as usually utilized for the Gabor edge filter as well as the collinearity. There is certainly possible to optimize the orientations as well as the following parameters further, yet this is out of the scope of the current work. The other important parameters are the width, shape, and length of the detected edge-filter via the Gabor filter. We choose a Gabor filter that detects a horizontal edge with a length of 11 pixels, a dark-bright shape and width of the edge of about 5 pixels (1/2 of 11 pixels), leading to a filter of $11 \times 11$ pixels kernel size, which is the default kernel size. A larger or smaller kernel size can adapt the filter kernel to broader or thinner edges, respectively, as well as it changes the length of the edge.

\subsection{Use case of SEM images domain}
\label{sec:resultsSem}
In the next section, we explore collinearity in a second use case, the recognition in scanning electron microscope (SEM) images. Within it, we will cover the saliency models and also deep learning based approaches.

\subsubsection{Data overview}

\begin{figure*}[t]
    \centering
    \includegraphics[width=1.0\textwidth]{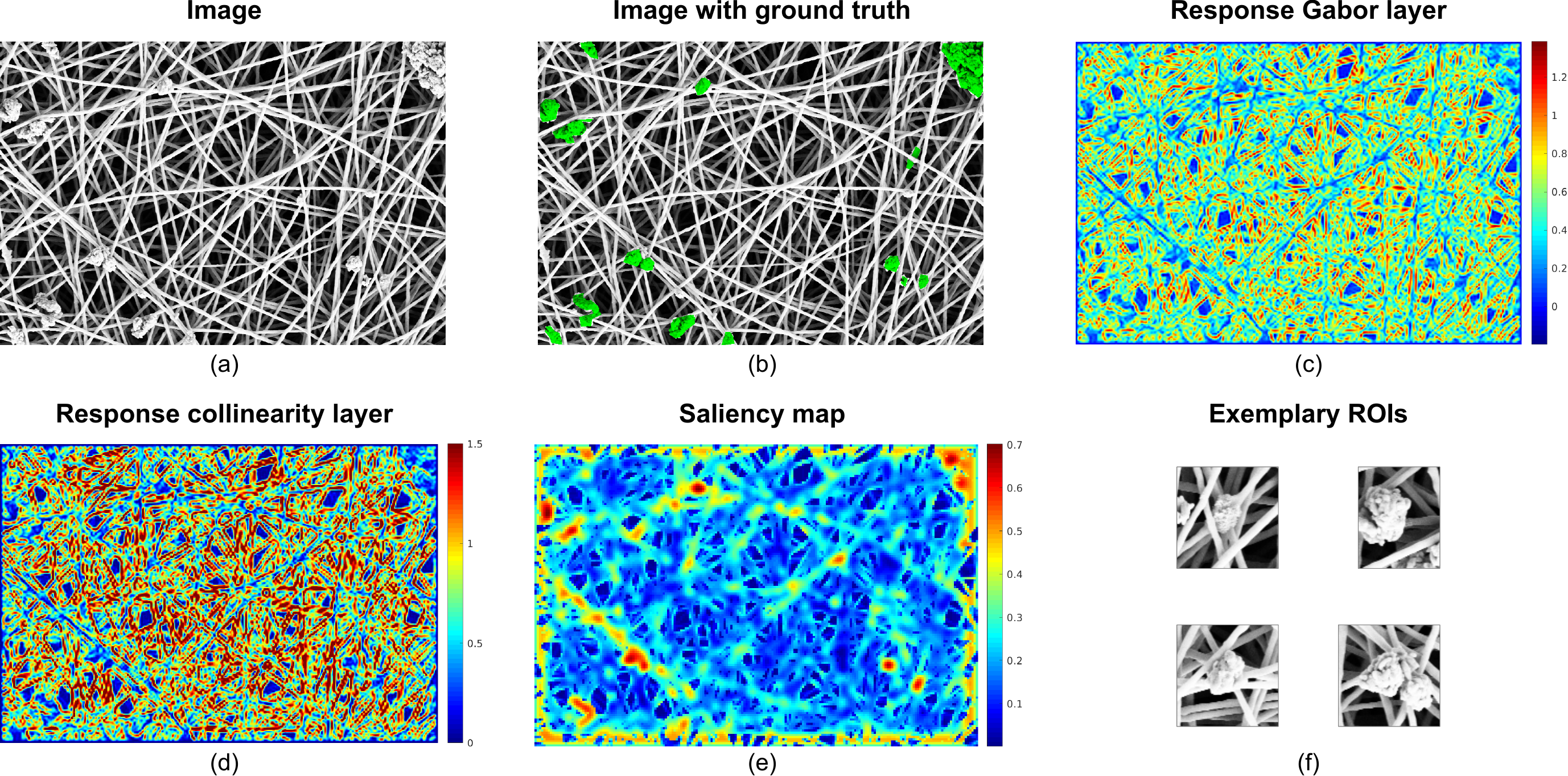}
    \caption{Saliency use case. \textbf{a, b)} Example image from the test set. The clumped structures denote the defects to be detected, for illustration-purposes shaded in green (b). \textbf{c)} Response of the Gabor layer. \textbf{d)} Response of the collinearity layer. For each response map, the values are illustrated as the maximum over all orientations. \textbf{e)} Saliency map. \textbf{f)} The first four extracted region-of-interests from the example image. }
    \vspace{-0.75em}
    \label{fig:resultsSEM}
\end{figure*}

As benchmark, we use a publicly available data set of industry material for defect detection. The data are images of nanofibrous materials taken via scanning electron microscopes (SEM) \citep{Carrera2017}. Our goal is to detect the clumped structures (defects) along the fibers emerging occasionally during the production process (Fig. \ref{fig:resultsSEM}a,b). The production is carried out via the the recent technique of electrospinning machines that directly produce woven nanofibrous materials \citep{Carrera2017}.
In the field of automatic inspection, data sets are often non-public due to industry, motivating us to use this publicly available data set. It is available for download at the following URL: \url{http://www.mi.imati.cnr.it/ettore/NanoTWICE/}.
The large-scale production of nanofibrous materials is indeed recognized as one of the main challenges in high-tech manufacturing \citep{Sengul2008,Meyer2009}, for example in the European Horizon 2020 program Factory of the Future. 

\subsubsection{Collinearity-based saliency models}
In this part, we explore the saliency models \citep{Itti1998,Itti2001,Borji2013}.
They use a “saliency stage” to detect region-of-interests (ROI), and subsequently, these ROIs are inspected by a classifier. Salient are conspicuous image structures or regions, and the idea reduces the necessary run time to process a scene. Conspicuous image structures are computed classically as red-green- and blue-yellow color contrasts, intensity, and edge orientation. An overview is found at \citep{Borji2013} and \citep[Chap. 2]{Beuth2019}. It has been a very common approach in the fields of biologically-inspired computational vision and is inspired by human visual attention. 
One of the goals of the saliency approach has been to reduce computational power, while this need has diminished today in the dawn of deep learning, yet low computational power is still in demand in industrial applications and low-power applications.
Since we found in our studies (cf. Sec. \ref{sec:resultsWafer}) that the collinearity stage alone as a classifier is not robust enough, we explore here the possibility to use the simple, biologically collinearity model as a saliency detector. Our finding is not surprising as collinearity is only one mechanism of the multiple mechanisms in the human visual system. In the human brain, the collinearity mechanism is part of the earlier processing, inspiring us to use it also in an earlier part of the processing chain and hence exploring saliency.

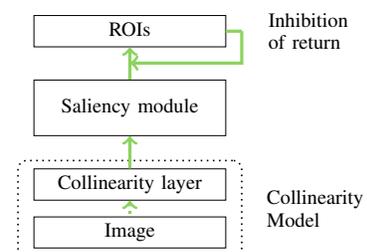
\begin{figure}[t]
	\centering
	\small
    \scalebox{0.85}{
	\begin{tikzpicture}[
        r/.style={draw, rectangle}]
    \node[r, minimum width=3cm, draw=none] (node5) at (1, 0.75) {Collinearity use case: SEM as a Saliency model};
    \node[r, minimum width=3cm,text depth=.2em] (node4) at (1, 0) {ROIs};
    \node[r, minimum width=3cm, minimum height=0.87cm,text depth=.2em] (node3) at (1, -1.2) {Saliency module};
    \node[r, minimum width=3cm,text depth=.2em] (node2) at (1, -2.4) {Collinearity layer};
    \node[r, minimum width=3cm,text depth=.2em] (node1) at (1, -3.15) {Image};
    \node[r, dotted, minimum width=3.5cm, minimum height=1.5cm, rounded corners, line width=0.25mm] (node0) at (1, -2.775) {};
    \node[r, draw=none, align=left] (node7) at (3.9, -2.775) {Collinearity \\ Model};
    \draw[<-, line width=0.5mm, arrowGreen] (node3) -- (node2);
    \draw[<-, line width=0.5mm, arrowGreen] (node4) -- (node3);
    \draw[<-, dotted, line width=0.5mm, arrowGreen] (node2) -- (node1);
    \node[r, draw=none, align=left] (node10) at (3.79, 0) {Inhibition \\ of return}; 
    \draw[<-, line width=0.5mm, arrowGreen] (1, -0.5) -| (2.75, 0);
    \draw[-, line width=0.5mm, arrowGreen] (2.75, 0) -| (2.5, 0);
	\end{tikzpicture}}
    \caption{Collinearity model in the use case as a saliency model. The notation of the figure is similar to Fig. \ref{fig:modelcaseFeaturedetector}. ROIs: Regions-of-interest.}
    \label{fig:modelcaseSaliency}
    \vspace*{-0.25em}
\end{figure}

\begin{table}[t]
  \caption{Results for saliency models - scanning electron microscopes (SEM).}
  \label{tab:resultsSEM_saliency}
  \centering
    \begin{tabularx}{0.95\columnwidth}{M{3.7cm}P{1.9cm}P{1.9cm}}
        \toprule
        & With collinearity     & Without collinearity \\
        \midrule
        ROIs Class ``Anomalous'' &  $~~\,313$   & $~~\,587$     \\
        ROIs Class ``Normal''    &  $~~\,804$   & $3\,814$  \\
        Total number of detected regions-of-interest (ROIs)  & $1\,117$ & $4\,401$  \\
        \bottomrule
    \end{tabularx}
\end{table}

Our proposed collinearity model is sketched in its saliency-modus-operandi in Fig. \ref{fig:modelcaseSaliency}. We process the images first with the collinearity model (Fig. \ref{fig:resultsSEM}a - d), whereas we use 32 different orientations. On top of its last neural layer, we extend the collinearity model with a saliency module, which produces a saliency map (Fig. \ref{fig:resultsSEM}e). In the current application, we look for the clumped structures constituting the defects, and the defects are non-collinear, while the flawless fibers are collinear (Fig. \ref{fig:resultsSEM}a). Clumpy defects are characterized by being not collinear, hence, we define everything as salient which “is not collinear” (Eq. \ref{eq:SEM_saliency1}, Eq. \ref{eq:SEM_saliency1b}), employing this as a major trick. Furthermore, we add a second constraint of "having a structure in the image” to ensure that empty regions are also not recognized as salient, because these regions a non-collinear too (Eq. \ref{eq:SEM_saliency2}). 
This result in the following saliency core code (Eq. \ref{eq:SEM_saliency1} - \ref{eq:SEM_saliency3}):
\vspace*{-0.5em}
\begin{eqnarray} 
  \label{eq:SEM_saliency1} S1_{x_1,\,x_2} &=& \max_{l'} \left(r\Area{Col}_{x_1,\,x_2,\,l'}\right) \\
  \label{eq:SEM_saliency1b} S_{x_1,\,x_2} &=& 1 - S1 \\
  \label{eq:SEM_saliency2} A_{x_1,\,x_2} &=& \left(r\Area{Img} \ge 0.5\right) \\
  \label{eq:SEM_saliency3} saliencyMap_{x_1,\,x_2} &=& \min\left(S_{x_1,\,x_2},\, A_{x_1,\,x_2}\right)
\end{eqnarray}
In them, $S$ denotes the saliency, the elementwise minimum of $S$ and $A$ constitutes the fuzzy-minimum operator, which is a logical “and” of two fuzzy sets, and is employed as by previous work \citep{Carpenter2003a}. To account for pixel inaccuracies, both maps are pooled in advance too. 
An inhibition of return approach (IOR) is implemented as commonly (\citep{Itti1998,Hamker2005b}), which leads to that attention is shifted to the next location and a new ROI is found in a loop, generating a list of regions-of-interest.
The classical saliency models contain particular features in several maps - typically color-contrasts, orientations, intensities, etc. We reinterprete this idea, and define as feature the collinearity, given by the collinearity layer.

As an ablation-study setup, we run the model without the collinearity stage. Hence, the map $S1$ and the saliency module are directly processed from the pool Gabor layer in this configuration:
\vspace*{-0.3em}
\begin{equation} 
  \label{eq:SEM_saliency4} S1_{baseline} = \max_{l'} \left(r\Area{Pooling}_{x_1,\,x_2,\,l'}\right) 
\end{equation}
\vspace*{-0.3em}

In our test environment, we found that the saliency model quantitatively produces -- over the whole data set of 40 images -- precisely 1117 region of interests (Tab. \ref{tab:resultsSEM_saliency}). Some of them are depicted in Fig. \ref{fig:resultsSEM}f. Of course, not all of them show defects, yet for this purpose a subsequent classifier detects them. We just conduct a count-based metric here, for a list of other possible metrics, please refer to \citep{Borji2013}.
In our ablation study setup, without collinearity, the number of extracted regions-of-interest increases to 4401. Therefore, the performance of the saliency model compared to a baseline approach is better by a factor of 3.9x. It implies that a subsequent classifier would need to investigate 3.9 times fewer regions-of-interest' images. This leads to a potentially higher detection rate and might enable to apply the system by a process engineer within low-power or low-computing devices within the production chain of the manufacturing process. The full results are provided and downloadable under \url{https://github.com/fbeuth/collinearity/tree/main/SEM/resultsSEMRois/}. 
By this study, we illustrate the way of how the collinearity principle might operate as a saliency model, as well explore the collinearity model's deploying in a second data set.

\textit{Discussion:} The classical saliency model(s) \citep{Itti1998} also incorporate center-surround structures, and incorporate a processing of the image at multiple-scales. We left out both operations to have a more straightforward approach, we feel that including them would make the performance just absolutely better on both the collinearity based experiment and the ablation study. Regarding center-surround structures, our structures of interest (edges and clumped non-edges) would not benefit much from center-surround processings. Both operations are different mechanisms and our focus is not to measure them, yet surely, they could be added in a future work. Newer saliency models also utilize learning, and we though of including it, yet then we would measure a mixed effect of collinearity and learning/deep learning, thus we also refrain to use this option to receive a clearer measurement result. The learning-based saliency models are often synonym with top-down saliency models \citep{Borji2013}. This is definitely worth to be also added in a future work.

\begin{figure}[t]
	\centering
    \small
    \hspace*{0.5cm}
    \scalebox{0.85}{
    \begin{tikzpicture}[
        r/.style={draw, rectangle}]
    \node[r, minimum width=3cm, draw=none] (node5) at (2.75, 0.75) {Deep learning in the SEM images domain};
    \node[r, minimum width=3cm, text depth=.2em] (node4) at (2.75, 0) {Class};
    \node[r, minimum width=3cm, minimum height=0.87cm,text depth=.2em] (node3) at (2.75, -0.95) {Deep neural network};
    \node[r, minimum width=2.5cm, text depth=.2em] (node2) at (1, -2.4) {Image};
    \node[r, minimum width=2.5cm, text depth=.2em] (node1) at (4.5, -2.4) {Feature detector};
    \node[r, minimum width=2.5cm, text depth=.2em] (node8) at (4.5, -3.35) {Collinearity layer};
    \node[r, minimum width=2.5cm, text depth=.2em] (node9) at (4.5, -4.1) {Image};
    \node[r, dotted, minimum width=3.0cm, minimum height=1.5cm, rounded corners, line width=0.25mm] (node0) at (4.5, -3.725) {};
    \node[r, draw=none, align=left] (node7) at (7.1, -3.725) {Collinearity \\ Model};    
    \node[r, draw=none, align=left] (node6) at (0.7, -1.85) {Channel 1};
    \node[r, draw=none, align=left] (node7) at (4.8, -1.85) {Channel 2};
    \draw[<-, line width=0.5mm, arrowGreen] (node3) -- (node1);
    \draw[<-, line width=0.5mm, arrowGreen] (node3) -- (node2);
    \draw[<-, line width=0.5mm, arrowGreen] (node4) -- (node3);
    \draw[<-, dotted, line width=0.5mm, arrowGreen] (node8) -- (node9);
    \draw[<-, line width=0.5mm, arrowGreen] (node1) -- (node8);
	\end{tikzpicture}}
	
    \caption{Collinearity model in the use case of deep learning in the SEM images domain. As feature detector, we utilize the saliency map from before (cf. Fig. \ref{fig:resultsSEM}e). The notation of the figure is similar to Fig. \ref{fig:modelcaseFeaturedetector}.}
    \label{fig:modelcaseDLSEM}
\end{figure}
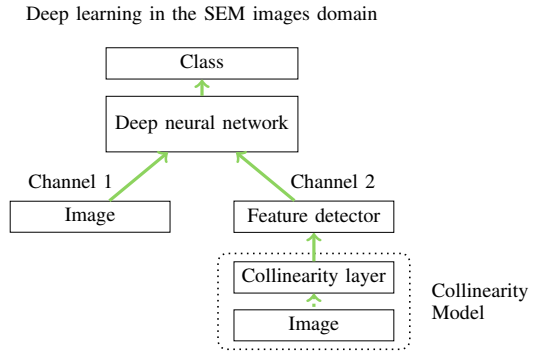

\subsubsection{Deep learning in the SEM images domain}
\label{sec:resultsSemDL}
As the next option, we illustrate and benchmark also deep learning \citep{LeCun2015} in combination with collinearity. The basic idea is to take advantage of collinearity as a feature detector for a subsequent deep neural network.  

As a feature detector, we build a detector alike to the previous section considering SEM. First, we create a feature detector that encodes features with the characteristic of ``it is not collinear'' for detecting the clumpy defects. The detector reacts to properties of pixels that have a collinearity layer response $r\Area{C}$ not much higher than their Gabor pooling response $r\Area{Pooling}$, which represents the “is not collinear” idea. In addition, the detector again reacts only to pixels that have image content to avoid a hit at empty image regions. For our detector, we re-purpose the same equations as in the previous section (Eq. \ref{eq:SEM_saliency1} - Eq. \ref{eq:SEM_saliency3}). After creating the feature detector, we feed the neural activity of the new feature detector together with the original grayscale image into a deep neural network (Fig. \ref{fig:modelcaseDLSEM}).

\begin{table}[t]
    \caption{Data set for deep learning in the SEM domain. Examples show the two classes of ``anomalous'' (defects) and ``normal''.}
    \label{tab:dataSEM_deeplearning}
    \centering
    \begin{minipage}{0.45\columnwidth}
       \centering
       \textbf{Class ``Anomalous''} \\
       $\estimates$ Non-Collinear \\
       \vspace{0.5em}
       \includegraphics[width=0.3\columnwidth]{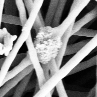}
       \vspace{1em}
    \end{minipage}
    \begin{minipage}{0.45\columnwidth}
       \centering
       \textbf{Class ``Normal''} \\
       $\estimates$ {Collinear} \\
       \vspace{0.5em}
       \includegraphics[width=0.3\columnwidth]{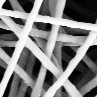}
       \vspace{1em}
    \end{minipage}

    \begin{tabularx}{\columnwidth}{X P{1.2cm}P{1.45cm}P{1.2cm}M{1.2cm}}
        \toprule
        Class & Train     & Validation & Test set & Total\\
        \midrule
        Anomalous & 265 & 36 & 37 & 338 \\
        Normal    & 299 & 33 & 18 & 350 \\
        \midrule
        Sum       & 564 & 69 & 55 & 688\\
        \bottomrule
    \end{tabularx}
\end{table}

We created a new data set from the original SEM data set by cutting out the detected faults along with many non-faults into small images, and by sorting the images into two classes: `anomalous' and `normal'. The original data set is pixelwise annotated and not suitable for a classification task via a DNN. The sorting is based on the ground truth of the original authors \citep{Carrera2017}, and our data set can be downloaded at \url{https://github.com/fbeuth/collinearity/tree/main/SEM/deepLearningSEMData/}.
Table \ref{tab:dataSEM_deeplearning} gives an overview of the data set. The entire data set was then randomly split into $80\,\%$ training, $10\,\%$ validation, and $10\,\%$ test set.

As a deep neural network, we utilize the following network as shown in Appendix Tab. 2, it is inspired by the VGG network \citep{Simonyan2015}. It consists of 3 VGG blocks, and contains $942\,882$ trainable parameters. We decide to rather use a small network to make the approach suitable for the mechanical engineering field since there is also a concern to use the approach for embedded GPU devices. Larger deep neural networks as a backbone would surely be possible. As methodological details, we employ the common class average accuracy as a metric for quantifying the classification performance: The accuracy is calculated separately for each class and then averaged across classes. We conduct 5 runs and all results are averaged over the runs.

As result, our dedicated test environment shows an achieved accuracy of $93.36\,\%$ as illustrated in Tab. \ref{tab:resultsSEM_deeplearning}. In an ablation study, we compare our system with a deep neural network without collinearity. For this purpose, we feed the dataset without the collinearity feature detector into the deep neural network, i.e. the DNN receives the raw image data of a grayscale image(s). Thus, the deep neural network has only one channel as input, not two, while we otherwise employ exactly the same neural network. We measure an increase in the accuracy from $78.35\,\%$ to $93.36\,\%$ by adding collinearity, hence an increase in accuracy of $15.01\,\%$ (Tab. \ref{tab:resultsSEM_deeplearning}). This implies that the error rate decreases from $21.65\,\%$ to $6.64\,\%$ when collinearity is added, which infers a performance improvement by a factor of $3.2\times$.

\begin{table}[t]
    \caption{SEM deep learning results. Accuracy in $\%$, along with the standard deviation. All the results are measured over five runs.}
    \label{tab:resultsSEM_deeplearning}
    \centering
    \begin{tabularx}{\columnwidth}{X P{2.2cm}P{2.2cm}}
        \toprule
        & With collinearity     & Without collinearity \\
        \midrule
        Custom CNN & $\mathbf{93.36 \pm 3.9}$ & $\mathbf{78.35 \pm 14.2}$  \\
        \midrule
        DenseNet121 \citep{Huang2017}   & $95.10 \pm 2.5$ & $89.20 \pm  3.9$  \\
        ResNet50 \citep{He2016a}      & $50.00 \pm 0.0$ & $50.50 \pm  6.3$  \\
        Xception \citep{Chollet2017}      & $93.10 \pm 5.7$ & $82.50 \pm  7.5$  \\
        \bottomrule
    \end{tabularx}
\end{table}

In addition, we benchmark some standard convolutional neural networks from the computer vision field (Tab. \ref{tab:resultsSEM_deeplearning}, DenseNet121 \citep{Huang2017}, ResNet50 \citep{He2016a}, Xception \citep{Chollet2017}). We observe accuracies of $95\,\%$ and $93\,\%$ for the  DenseNet and Xception networks respectively, and without collinearity, accuracies of $89\,\%$ and $82\,\%$ respectively. This denotes an increase of $6\,\%$ and $11\,\%$. The error rates decrease from $11\,\%$ to $5\,\%$ and from $18\,\%$ to $7\,\%$ respectively, implying an improvement by factors of $2.2\times$ and $2.6\times$. Hence, we found a similar pattern of results that the accuracy increases through the collinearity method. ResNet50 does not learn the task, both with and without collinearity, and we suspect that the low amount of data (cf. Tab. \ref{tab:dataSEM_deeplearning}) is the cause as ResNets are known notoriously for requiring a large amount of data. Note, a deeper analysis reveals that these larger networks have a strong overfitting, thus we do not consider their accuracy results here as the best option, and we hypothesize that the overfitting is caused by the extremely small data set (Tab. \ref{tab:dataSEM_deeplearning}). For larger datasets, the larger networks might perform better.
Therefore, the feature detector by collinearity seems to serve as a second source of information for the deep neural network(s), and hence strongly benefiting quantitatively the recognition accuracy up to a factor of 3.2x and of 2.6x respectively. To summarize, here, we illustrated as well as benchmarked deep learning with collinearity on a second use case.

\begin{figure}[t]
    \centering
    \subcaptionbox{~}{\includegraphics[height=3.25cm]{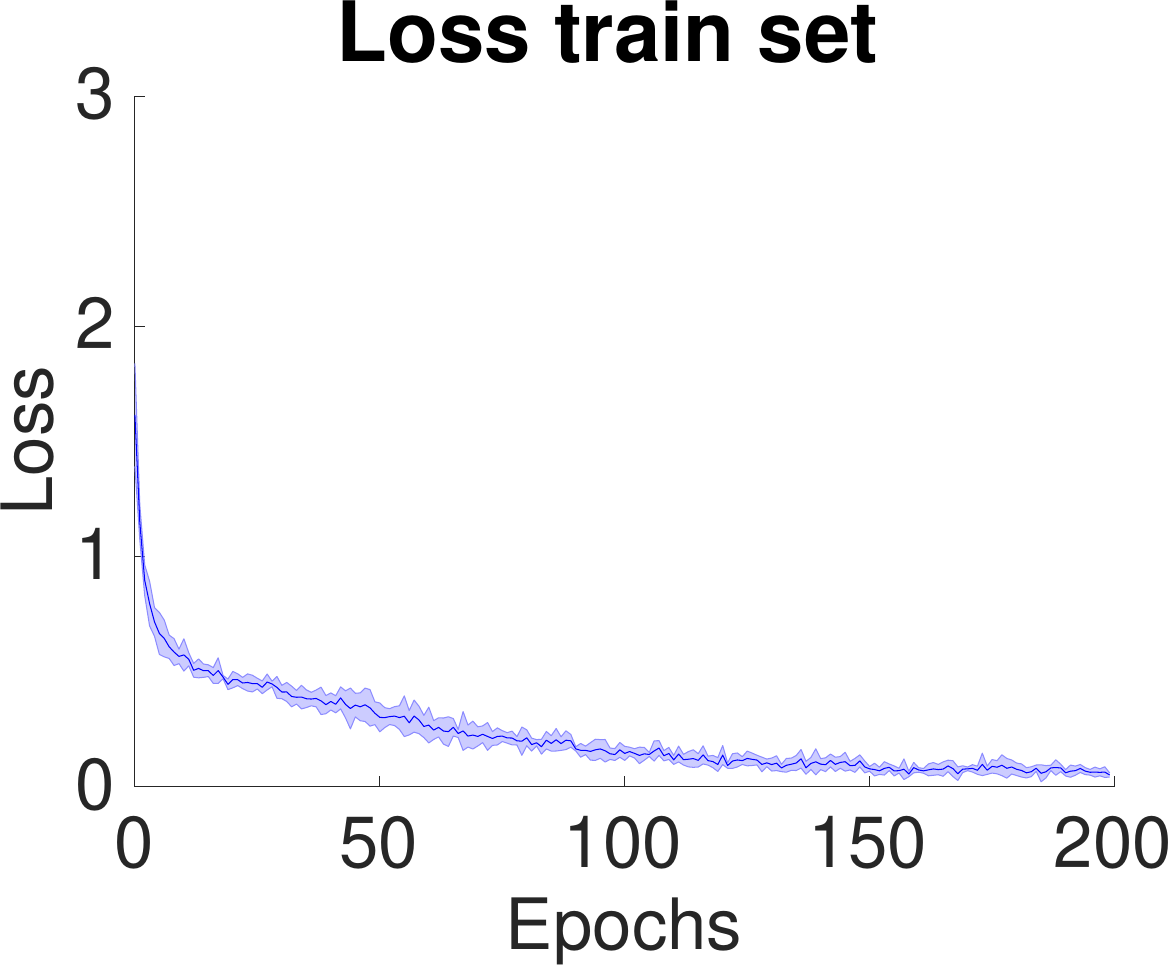}}%
    \hspace{0.4cm}
    \subcaptionbox{~}{\includegraphics[height=3.25cm]{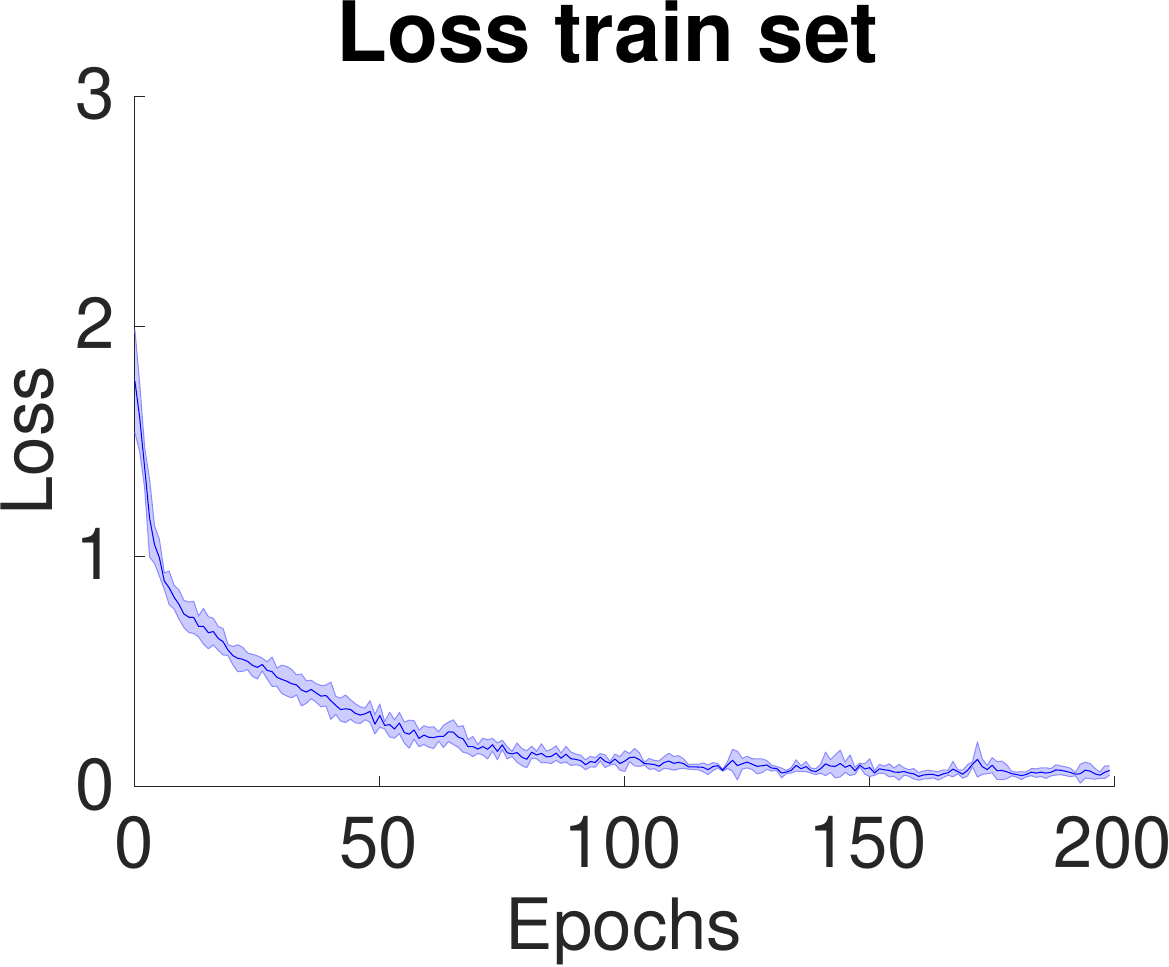}}\vspace*{0.3cm}%
    \\
    \hspace{0.05cm}
    \subcaptionbox{~}{\includegraphics[height=3.25cm]{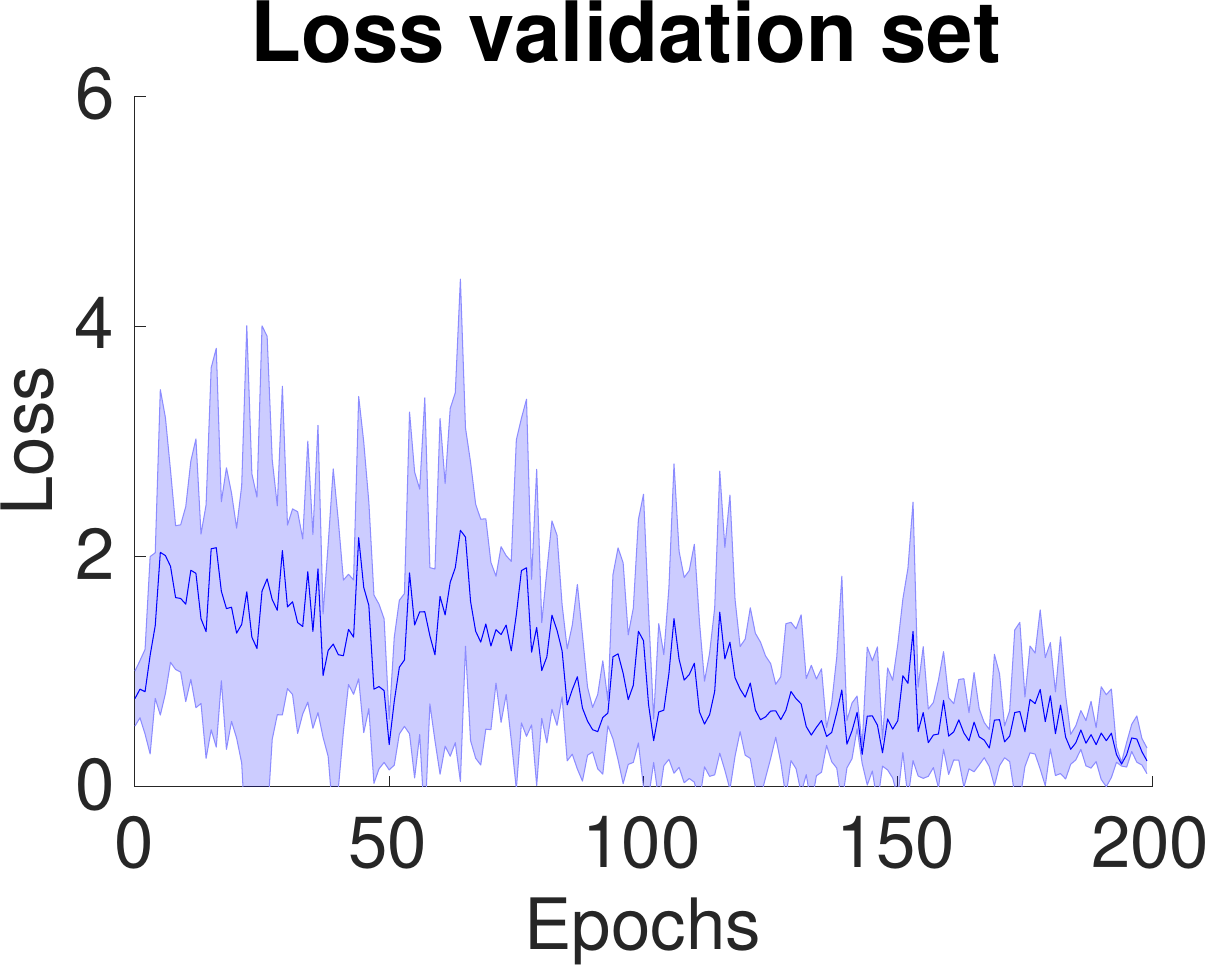}}%
    \hspace{0.25cm}
    \subcaptionbox{~}{\includegraphics[height=3.25cm]{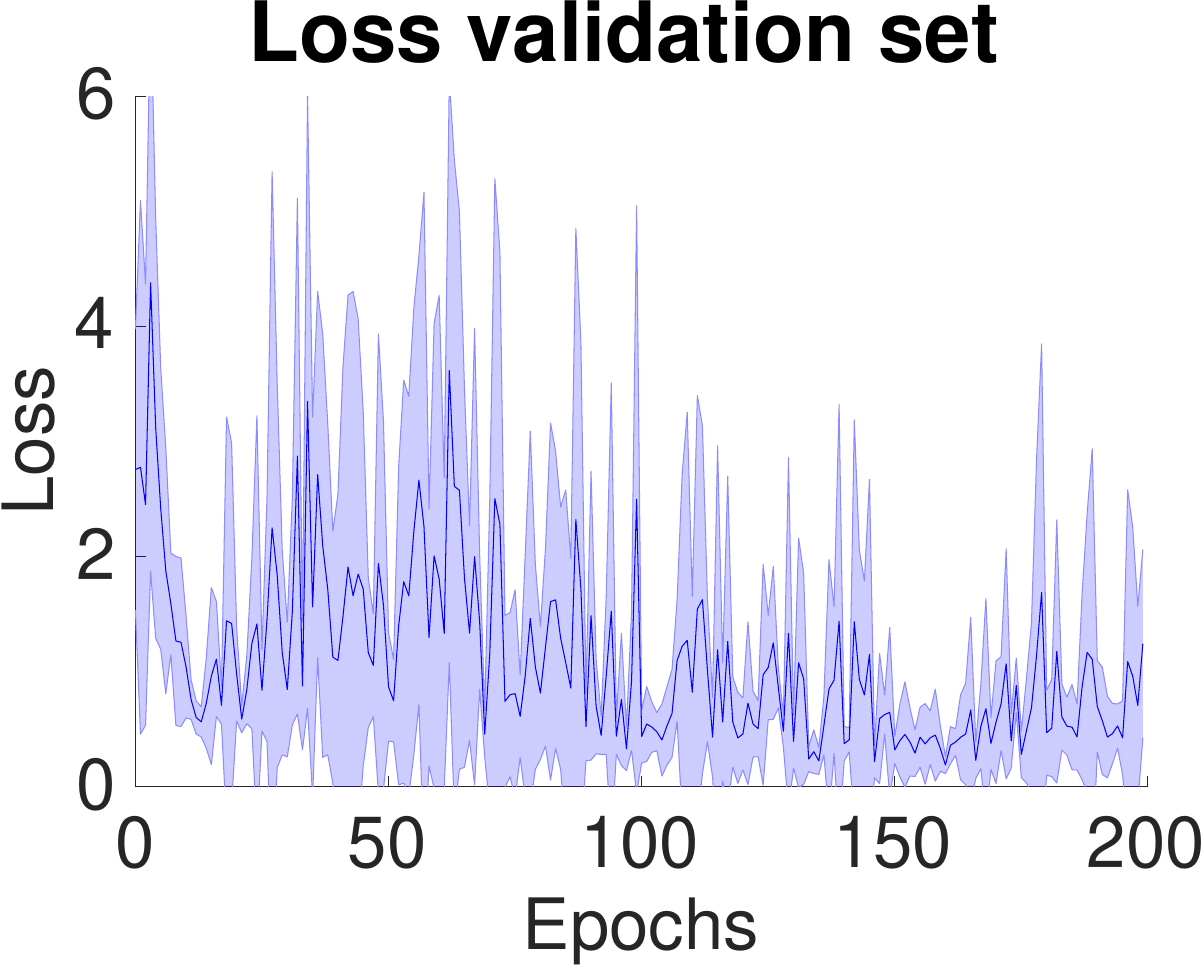}}%
    \caption{Deep learning stability in the SEM use case. \textbf{a, c)} With collinearity. \textbf{b, d)} Without collinearity. It is analyzed over multiple runs and shows the mean of the multiple runs' loss metrics (solid line) as well as the standard deviation (shaded area).}
    \label{fig:resultsSEM_dlStability}
    \vspace*{-1em}
\end{figure}

\textit{Improved deep learning stability}:
In addition, we observe that the training process of the deep neural network is notable more stable with collinearity. 
This finding indicates a higher reliability of the training, and, therefore, it saves time and labor for the analyst and reduces costs. We analyze our finding quantitatively by evaluating the loss metric over time across multiple runs, and calculating the standard deviation of the loss across runs. The basic idea is that a lower standard deviation across runs implies a more reliable training since all runs behave in a similar manner. The standard deviation, along with the mean, is plotted as shaded error bar, thus a lower standard deviation is characterized by a smaller shaded area and shows less variance between the runs (Fig. \ref{fig:resultsSEM_dlStability}).

For the training metrics, we observe a slightly faster decrease of the loss with collinearity (Fig. \ref{fig:resultsSEM_dlStability}a and \ref{fig:resultsSEM_dlStability}b), which indicates that the training proceeds faster. Yet, the effect is small for the training loss and is mostly visible between the 2nd and 30th epoch. It is more strongly observable in the validation loss, where over all epochs, the mean loss decreases faster with collinearity than without collinearity (Fig. \ref{fig:resultsSEM_dlStability}c and \ref{fig:resultsSEM_dlStability}d, solid line). Although the amount of the training speedup is low to average, the effect is consistent.

Much more interesting is the finding that the training process is more stable with collinearity. 
Fig. \ref{fig:resultsSEM_dlStability}c, which includes collinearity, shows less fluctuation compared to Fig. \ref{fig:resultsSEM_dlStability}d without collinearity, indicating that the runs operate more similarly and less varying, and thus a more stable training process with collinearity. This is seen by the lower standard deviation between the runs. Hence, the training procedure is more reliable. 
Furthermore, we notice without collinearity a slight tendency of overfitting, which is visible after epoch $160$ when the validation loss increases while the training loss remains low (Fig. \ref{fig:resultsSEM_dlStability}b and \ref{fig:resultsSEM_dlStability}d). To summarize, we found that collinearity increases the tendency of the training procedure to operate more stably and reliably which is observed by less variance between runs, a slightly faster training, and avoiding a tendency of overfitting.

\subsection{Use case of interruptions and occlusions}
\label{sec:resultsOcclusion}

\begin{figure}
    \centering
    \includegraphics[width=1.0\columnwidth]{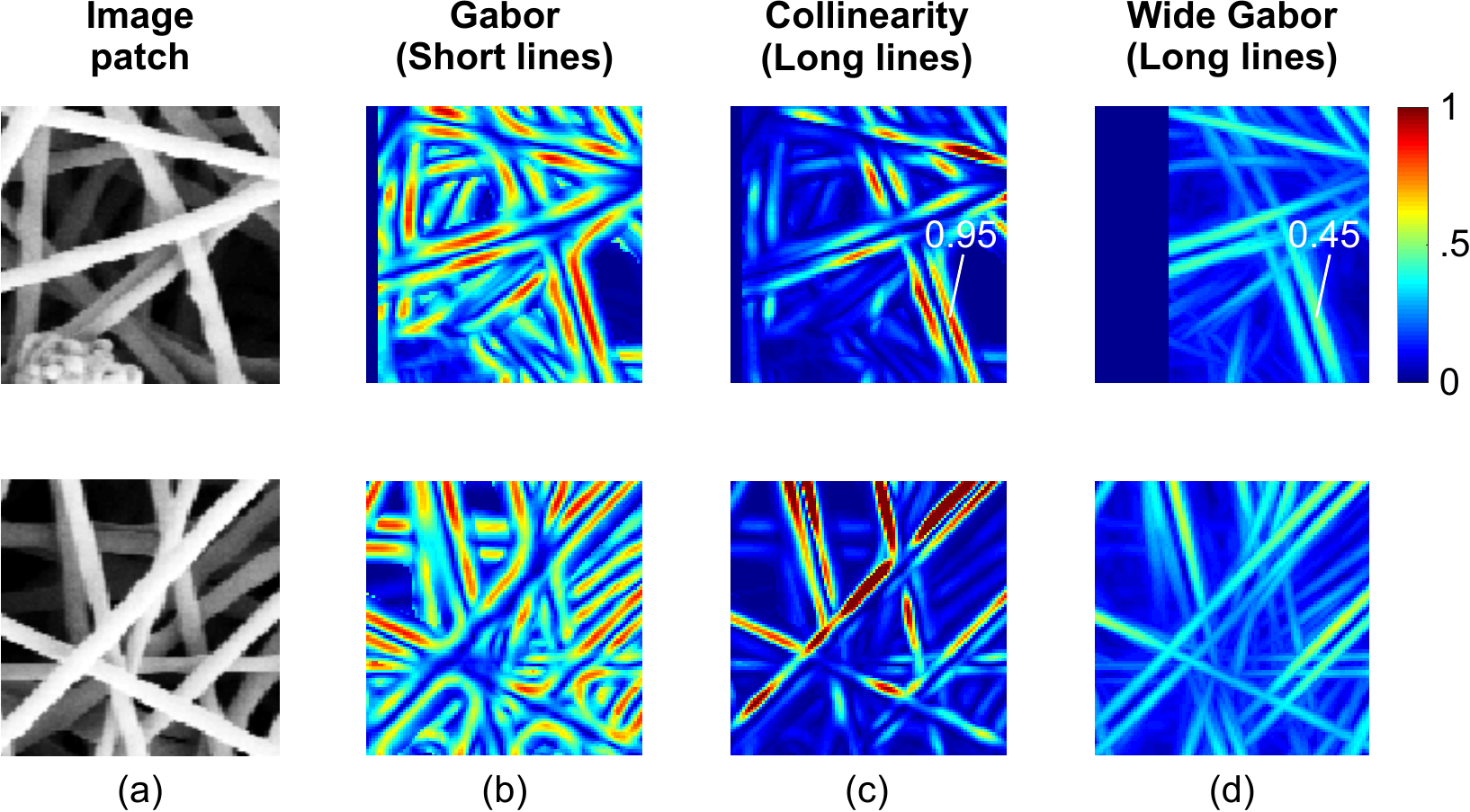}  
    \caption{Interruptions and occlusions. \textbf{a)} Two exemplary image patches from SEM with interrupted and occluded lines. \textbf{b)} Gabor response. The response is shown as an overlay over all orientations, obtained by a maximum of the responses over all orientations. \textbf{c)} Collinearity response. \textbf{d)} Response when utilizing a broad Gabor filter to detect the long lines, here chosen as $5\times$ the size of the Gabor in (b). (d) is an ablation study. Both (c) and (d) are options for detecting longer lines.}
    \label{fig:resultsOcclusion}
\end{figure}

As third case, we investigate the usage of collinearity for the interruptions and occlusions of objects. As opportunity, we use the SEM data of the previous section since the data set contains many fibers (lines) which are overlaid by other fibers. Hence, the fibers are occluding or interrupting each other, which enables us the study (Fig. \ref{fig:resultsOcclusion}a). We first observe that the Gabor filters react to the lines' borders, and also to shorter line segments, therefore they encode “short lines” (Fig. \ref{fig:resultsOcclusion}b). 
When we study the collinearity, we found that it reacts to longer lines (Fig. \ref{fig:resultsOcclusion}c at the top row, and even more prominent at the bottom row). 
Hence, we conclude that collinearity is an approach for encoding longer lines since its properties react to these objects in an image. These longer lines might be easily partially-occluded. The collinearity principle encodes the longer, partially-occluded lines by employing connected line-detectors that react to short segments of the structure, and collinearity then links these neural responses together. In this way, the occlusions are to a much lower degree hampering the recognition of long, occluded lines.

\begin{figure*}
    \centering
    \includegraphics[width=1.0\textwidth]{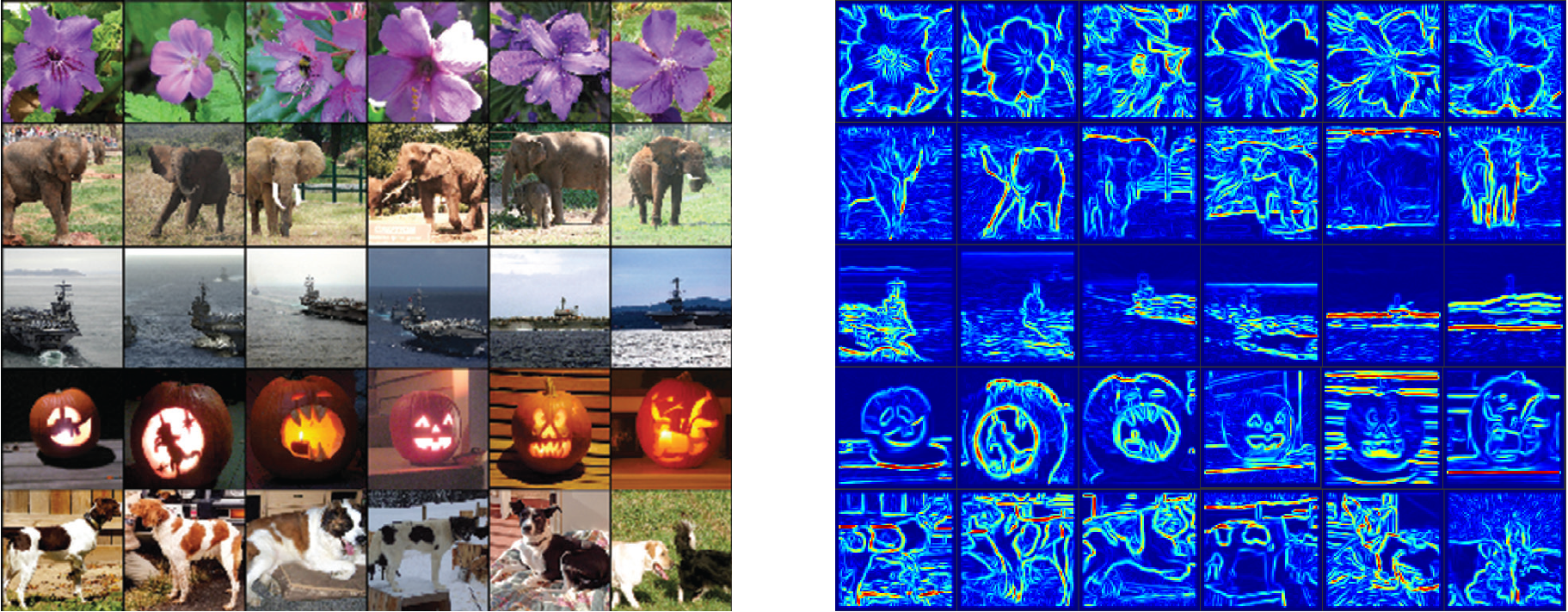}
    \caption{ImageNet. Taken from Krizhevsky et al. (2017)\citep{Krizhevsky2017}. Left are shown some test images and right the collinearity responses.}
    \label{fig:resultImagenet}
    \vspace*{-0.25em}
\end{figure*}

To determine the performance of the collinearity based approach to a standard method, we compare collinearity to a method where we extend the Gabor filters (Fig. \ref{fig:resultsOcclusion}d, ablation study). We expanded the Gabor filter from a $11\times11$ pixels kernel size to a $55\times55$ pixels kernel size, along with an appropriate expansion of $\sigma$, denoted “Long-Gabor”. The shown, cropped images are $96\times96$ pixels in size.
We observe the following outcomes: i) The collinearity approach enhances the long lines, while keeping the background activity low. Collinearity has a better signal-to-noise ratio because the noise in the Long-Gabor variant is higher (Fig. \ref{fig:resultsOcclusion}d). This originates from the fact that the kernels of the long-Gabor are large and they thus react often also to some other image elements. ii) Linked to this, the Long-Gabors have lower responses (in white) as due to the longer kernels they almost never reach optimal responses since they are occluded by lines and thus partially match.
iii) Both methods work for detecting longer edges. iv) Finally, the Long-Gabors are much more computationally expensive. They use a larger kernel of $55\times55$ which is also needed e.g. for Gabors with skewed orientations such as $45^\circ$, hence the complexity increases by O($n^2$). 

\textit{Discussion:} The current work addresses collinearity at the level of lines. From the architecture in the human brain, we know that lines are encoded by the first visual area, the primate visual cortex \citep{Hubel1968}. Higher visual cortices  encode objects (inferior temporal cortex \citep{Logothetis1995}). It is valid to assume that also the inferior temporal cortex has lateral connections, and we could assume that also objects can be addressed by collinearity, i.e. if the objects are arranged in a row. The proposed model could be surely appropriately extended to facilitate also collinearity via objects by: We propose to change the model's Gabor filter to a deep neural network model that is able to recognizes objects, and then applying collinearity on top of it. This idea would allow to utilize collinearity also with objects.

\newpage
Therefore, we conclude that the collinearity principle is usable for the detection of partially-occluded structures in images or interrupted lines. Occlusions are an important topic in many image applications such as automotive \citep{Gilroy2021} where the challenge is to detect partially-occluded pedestrians in streets. Maybe collinearity contributes to the brain mechanisms to deal with the occlusion problem.

\subsection{Use case of real-world data at ImageNet}
\label{sec:resultsImagenet}

We also evaluate ImageNet \citep{Krizhevsky2012} with our collinearity method.
ImageNet is a well-known challenge to benchmark models at natural scenes and objects, and has been used to benchmark many important deep learning models \citep{Krizhevsky2017,Russakovsky2015,Krizhevsky2012}.
We run our model on test images (Fig. \ref{fig:resultImagenet}a) with standard parameters (32 orientations, kernelsize = 11 pixels, $w\Area{Col} = 2$), and conduct a qualitative analysis. 
The analysis is evaluated regarding the neural collinearity response for the test images (Fig. \ref{fig:resultImagenet}b). In the  illustration, the collinearity responses are shown as the maximum over all 32 orientations. 
We observe on our exemplary test images that the collinearity response represents the image data not very well. The reason behind this fact is that many of the ImageNet classes do not consist of lines, but rather more of natural-grown objects. For example, we found that the class elephant (second row, Fig. \ref{fig:resultImagenet}b) and the class dog (fifth row) are not well presented as visible from the neuronal responses. Other classes such as the aircraft carrier (third row) and pumpkin (fourth row) are better represented. We observe surprisingly also the category flower (first row) is well represented because the flowers' petals consist of long line elements. In contrast, the dogs' inner structure -- its texture -- is completely irregular, and collinearity cannot benefit in this class much.
These observations let us conclude that the collinearity method might not be so well-suited for natural-grown objects, first due to irregular contours and secondly due to irregular inner textures of an object. Of course, the suitability depends in detail on the particular class. In addition, it lets us infer that the collinearity principle might not be so valuable for existing data sets, which are often derivations from or are similar to ImageNet, hence, the method might also not show its power until now.

\section{Strengths and Limitations}
\label{sec:limits}

We have proposed in this work a model for exploiting long-line structures in an image (collinearity). The model is based on the primate early visual system. Despite the model’s simplicity, and that it is based as the primary component on a long kernel, the model has a strong performance to distinguish collinear image structures from not-collinear ones. In an appropriate image environment, the approach increases performance by magnitudes, e.g. up to a factor of 3.2x in the SEM image domain and up to 1.2x in the fault detection of semiconductor wafers.

Despite the understanding of the human and primate visual system, not much has been known about this principle for which purpose it is used in real-world environments and applications, hence, we could shed light on its usage: We found that it is mainly used to enhance the contrast when particular-structured stimuli occur. Such properties of a structure could be a distinguishable characteristic between classes, and thus benefit recognition as well as detection accuracies. Our goal here is to transfer this computing principle to real-world domains and make it useful for computer vision and engineering applications, and in the long term, to facilitate the power of the human visual system.

A good performance can be expected from the model when the objects of interest are collinear, or when the classes differ in the collinearity property yet other features are the same. This is what we found, it is not solely the collinear characteristic that helps, rather it is that collinearity is the differing property. For example, the approach is capable of detecting the wafer faults, because they are not-collinear while the other class (not-faulty) is collinear (Fig. \ref{fig:resultsWafer} top). Otherwise, both classes have a similar appearance (e.g. lines, textures) and color (gray). Similarly, in the SEM data use case, the targeting faults are clumpy structures, but they have a good amount of variability in size, texture, and shape (high inter-class variability). Likewise, the non-faults are fibers, but with all shapes, sizes, occlusions, and orientations. The main distinguishing property between the two classes of faults and non-faults is collinearity (Tab. \ref{tab:dataSEM_deeplearning}).

Not implemented in the model are reactions to curves. The primate brain contains also cells reacting to curved patterns \citep{Kapadia1995} as well as shown also by the learning from statistics of natural scenes \citep{Hoyer2002, Prodohl2003}. A deeper analysis from \citep{Kapadia1995} reveals that divergent cell populations are utilized for straight lines and other populations for curves, respectively. Hence, if we would like to implement curved patterns, we would need to double at least the number of cells, which would make the model slower and more complex. Thus, we decide against implementing curves. One could incorporate curves by changing the pattern of the collinearity connectivity matrix, i.e. connecting cells in the form of the desired curve ($w_{x'_1,x'_2}$, see Eq. \ref{eq:col3}).

Similarly, the model does not cover the detection of structures arranged in more complex patterns, which can only be learned. Learning to facilitate complex patterns is performed via the learning of second-order statistics from natural scenes \citep{Hoyer2002}. Introducing an extra learning step would might make the model less appealing to users, thus, we leave it out, and the learning is discussed somewhere else \citep{Hoyer2002, Prodohl2003}. Learning could be introduced by an appropriate learning step, and then by exchanging the connectivity matrix ($w_{x'_1,x'_2}$).

Finally, there exist many other properties of collinearity (Polat \& Sagi (1993, 1994) \citep{Polat1993,Polat1994a}; Bonneh \& Sagi (1998) \citep{Bonneh1998}; Shani \& Sagi (2005) \citep{Shani2005}; Polat (2009) \citep{Polat2009}; Chan et al. (2012) \citep{Chan2012}; and Maniglia et al. (2015a, 2015b, 2022) \citep{Maniglia2015a, Maniglia2015b, Maniglia2022}). 
They are not included and might be covered by future work. For example, collinearity has a contrast dependent property, i.e. it reacts differently at high and low contrasts. Simulating such contrast property of cells is usually performed with the concept of a divisive-normalization activation function \citep{Albrecht1982, Carandini2012}, which would require of changing/introducing the internal activation function of the neurons, converting from currently a simple half-rectification to a logarithmic version with several parameters. Divisive normalization requires more experience with the concept, has its own additional parameters, and requires calibration. Thus, we found that it is not very user friendly, and we did not model this further, as it is probably not very often needed as well as it makes the neural model more complex. An adaptation of the activation function would be necessary to use it as outlined in our previous work \citep{Beuth2015a}. Therefore, the model has no special properties for dealing with low contrasts, while the human system has. Other properties of collinearity surely exist, which would require to model more details from primate and human neuroscience.

\section{Conclusion}
\label{sec:conclusion}

Collinearity is one of the biological processing principles in the visual system of humans. We explore in this work the principle to better understand its function in reality, and its usefulness for computer vision applications which is a largely unexplored field, linked to our goal of transferring collinearity processing to the computer vision domain. The collinearity principle is a concept to enhance and better perceive edges that are arranged in a straight line. Thus, we observe that the collinearity constitutes a structure-dependent image enhancement because it enhances the image dependent on the overall image structures present (Sec. \ref{sec:resultsRotation}). In our work, we first analyzed the concept systemically (Sec. \ref{sec:resultsSystematic}), replicated psychological data (Sec. \ref{sec:resultsSystematic}), and then explored its usefulness for computer vision applications (Sec. \ref{sec:resultsRealworld}).

We explore in this work collinearity in several computer vision application domains. By these means, we generate a list of use cases for which the collinearity principle is useful, and for which it is not promising.
In use case 1, we found that collinearity worked well for improving the fault recognition of semiconductor wafers. We build a fault recognition detector based on collinearity, and combine collinearity with deep neural networks, where it enhances the performance by a factor of $1.2$x (via an error rate decrease from $6.5$\,\% to $5.3$\,\%). In use case 2, we studied the recognition of fibers in nanotechnology materials, and combine collinearity with deep learning and investigate saliency models. In the deep learning case, collinearity notably improves the performance by decreasing the error rate from $21.6\,\%$ to $6.6\,\%$ (factor $3.2$x).
In the third use case, we investigated the enhanced recognition of occlusions (Sec. \ref{sec:resultsOcclusion}); while, in the fourth use case, we applied collinearity briefly to ImageNet and found out that it is not beneficial (Sec. \ref{sec:resultsImagenet}). The use cases are selected to illustrate the way to combine collinearity with deep learning (use cases 1, 2), show its usage as a feature detector intended for low-complex applications (use case 1), and sketch its deployment as a saliency model (use case 2). 

From our explorations, we found out that it is important that collinearity serves as a property to differentiate between classes, where one class is collinear and the other class is not collinear (Sec. \ref{sec:limits}, \ref{sec:resultsRealworld}), leading to a good performance. Secondly, the list of use cases let us infer that the principle might be well suited for industry applications since the industry has often image structures of interest that are “line-like”. Since industry has also often a low number of samples and collinearity does not require a lot of samples unlike deep learning, it might be very suitable for the recognition of patterns in man-made objects and industry applications.

Looking ahead, future work might develop the system further to facilitate also circular shapes. The human brain contains also such kind of cells \citep{Kapadia1995}, yet, these are not part of the proposed model and we plan them for a future extension of the model. Moreover, future work could investigate other use cases as more application domains with line- and man-made structures are surely existing in the industry.

\section*{Acknowledgements}
We would like to thank Waseem Farooq, who worked during his master thesis on a first version of the model and helped us to start the project. Furthermore, we thank Michael Teichmann for his co-supervision of said master thesis, his development of the evaluations in the early stage of the project, and his many fruitful discussions, and we thank Shabnam Novin for her comments and fruitful discussions on the figures and the manuscript text. This project was funded privately by F.B..

\newpage
\small
\bibliographystyle{IEEEtran}
\bibliography{collinearityBib_noMonth.bib}

\section{Biography Section}
%
%
\vspace{-33pt}
\begin{IEEEbiography}[{\includegraphics[width=1in,height=1.25in,clip,keepaspectratio]{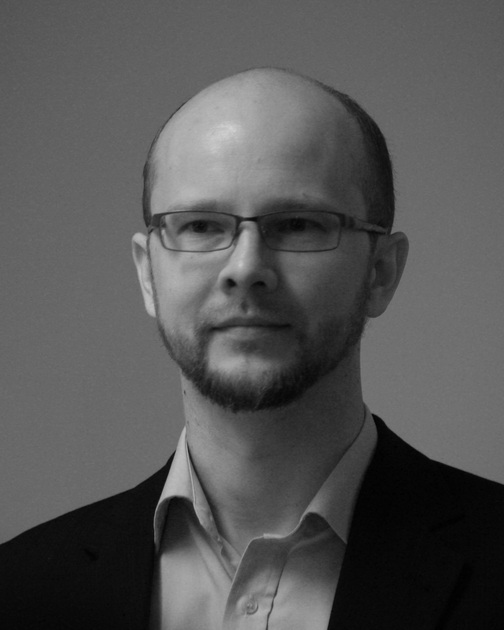}}]{Frederik Beuth}
received the Dipl.-Inf. degree in computer science and artificial intelligence from the Chemnitz University of Technology, Germany, in 2008. He did a doctoral degree under professor Hamker in artificial intelligence and computational-neuroscience, obtaining his degree from Chemnitz University of Technology, Germany, in 2018. His main focus lies in the topics of biologically-plausible models of the human visual system and visual attention, and related topics such as computer vision and deep learning. Currently, he is working at bringing neuro-computational operation principles and models towards real-world computer vision tasks. This line of work arises from the current affiliation in the team of Junior Professorship Media Computing, which conducts research in deep learning and computer vision for small and medium-sized enterprises, bringing a unique opportunity to work with neuroscientific principles and models at real-world computer vision problems in a flexible environment.
\end{IEEEbiography}

\vspace{-33pt}
\vspace{0.4em}
\begin{IEEEbiography}[{\includegraphics[width=1in,height=1.25in,clip,keepaspectratio]{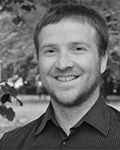}}]{Danny Kowerko}
was born in 1980 in Chemnitz, Germany. He received his diploma in physics and his Ph.D. degree at the department of Natural Sciences of the Chemnitz University of Technology, in 2005 and 2010, respectively. Till then, his research was dedicated to studies of energy transfer processes between semiconductor nanocrystals and organic dye molecules at the ensemble and single molecule level. As Postdoc at the University of Zurich in Switzerland, from 2011--2015, he focused on the computational analysis of RNA (ribonucleic acid) folding, conducted by single molecule and ensemble microscopic and spectroscopic methods. In 2015 he returned to the Chemnitz University of Technology and became junior professor at the department of computer science in 2019. His junior professorship “Media Computing” is dedicated to algorithm and software development in the field of computer vision, such as biocomputation, medical informatics, computational ophthalmology, automatic fault detection in semiconductor wafers, hexagonal image processing, and general object detection and classification.
\end{IEEEbiography}

\vfill

\end{document}